\pdfoutput=1
\documentclass[11pt]{article}

\usepackage[preprint]{acl}

\usepackage{times}
\usepackage{latexsym}
\usepackage{amsmath}

\usepackage[T1]{fontenc}
\usepackage[utf8]{inputenc}

\usepackage{microtype}

\usepackage{inconsolata}

\usepackage{enumitem}
\usepackage{graphicx}
\usepackage{caption}
\usepackage{multicol}
\usepackage{multirow}
\usepackage{diagbox}
\usepackage{makecell}
\usepackage{CJKutf8}
\usepackage{booktabs}
\usepackage{array}
\usepackage{supertabular}
\usepackage{longtable}

\newcommand{\PipeName}{CAT-BEAR}
\newcommand{\DatasetName}{CAPE}
\newcommand{\ModelName}{ChatEMO}
\newcommand{\MetricsName}{CAT-Dist}

\newcommand\blfootnote[1]{%
  \begingroup
  \renewcommand\thefootnote{}\footnote{#1}%
  \addtocounter{footnote}{-1}%
  \endgroup
}

\title{CAPE: A Chinese Dataset for Appraisal-based Emotional Generation using Large Language Models}

\author{
    June M. Liu\textsuperscript{1,2{*$\dagger$}} \quad  
    He Cao\textsuperscript{1,3{*$\dagger$}} \quad  
    Renliang Sun\textsuperscript{1,4*} \quad  
    Rui Wang\textsuperscript{1} \quad  
    Yu Li\textsuperscript{1} \quad  
    Jiaxing Zhang\textsuperscript{1} \\
    \textsuperscript{1} International Digital Economy Academy (IDEA),  
    \textsuperscript{2} University of Michigan, Ann Arbor \\  
    \textsuperscript{3} Hong Kong University of Science and Technology  
    \textsuperscript{4} University of California, Los Angeles \\  
    \texttt{juneliu@umich.edu},  
    \texttt{hcaoaf@connect.ust.hk},  
    \texttt{sunrenliang@ucla.edu}, \\  
    \texttt{\{wangrui,liyu,zhangjiaxing\}@idea.edu.cn}
}

\begin{document}
\maketitle

\begin{abstract}
Generating emotionally appropriate responses in conversations with large language models presents a significant challenge due to the complexities of human emotions and cognitive processes, which remain largely underexplored in their critical role in social interactions. In this study, we introduce a two-stage automatic data generation framework to create \DatasetName{}, a Chinese dataset named \textbf{C}ognitive \textbf{Ap}praisal theory-based \textbf{E}motional corpus. This corpus facilitates the generation of dialogues with contextually appropriate emotional responses by accounting for diverse personal and situational factors. We propose two tasks utilizing this dataset: emotion prediction and next utterance prediction. Both automated and human evaluations demonstrate that agents trained on our dataset can deliver responses that are more aligned with human emotional expressions. Our study shows the potential for advancing emotional expression in conversational agents, paving the way for more nuanced and meaningful human-computer interactions.
\end{abstract}

\blfootnote{$*$ Equal contribution. $\dagger$ Work is done during their internship at IDEA.}

\section{Introduction}
%（考虑了appraisal，以及一点点场景，主要是和记忆相关的过去发生的事情）\cite{croissant2024appraisal}

\begin{figure}[t]
\scalebox{0.95}{
\includegraphics[width=\linewidth]{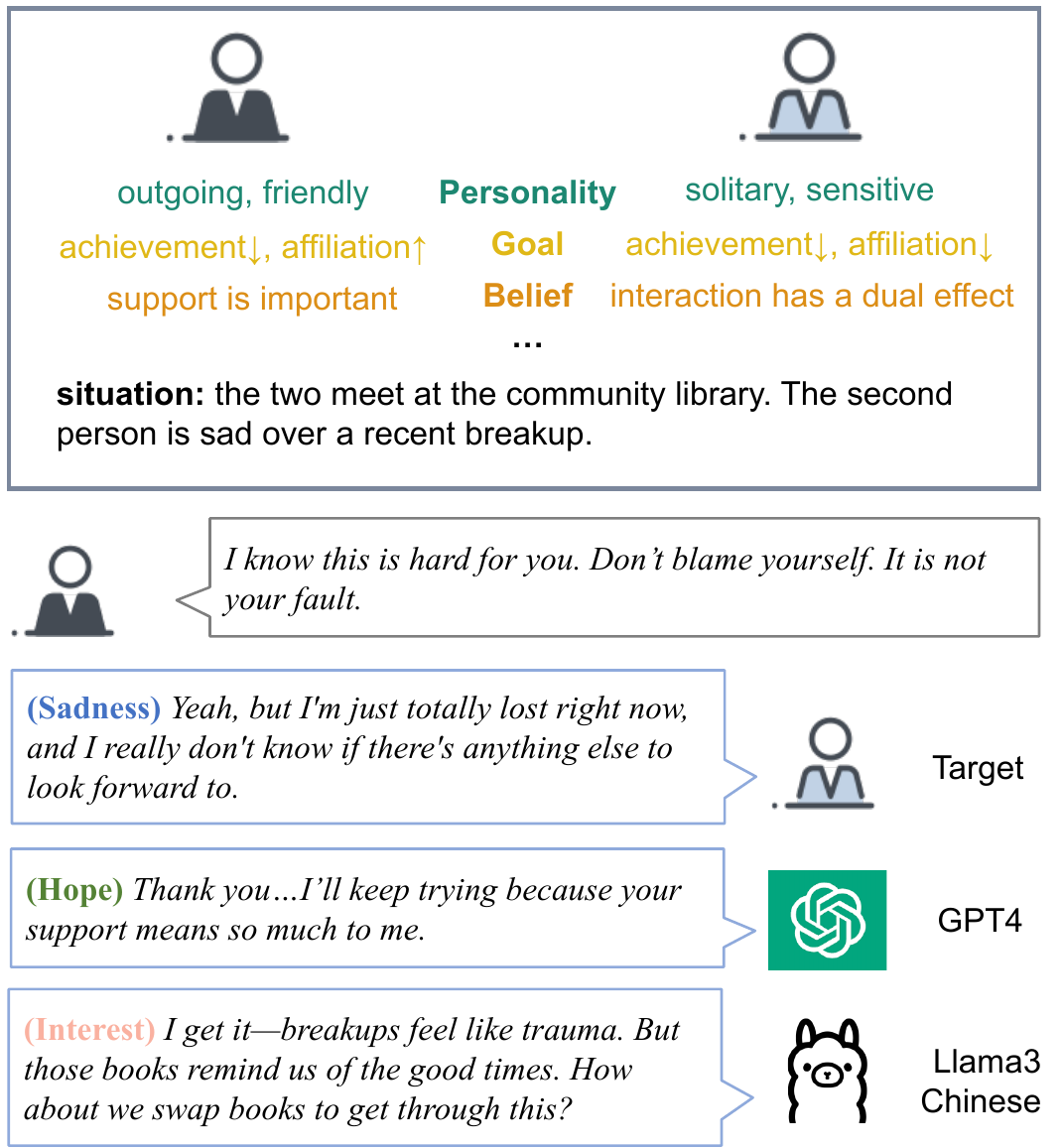}}
  \caption {An example of an emotion generation conversation with the analysis on the results of LLMs. }
  \label{fig:comparison}
\end{figure}

Emotion is a crucial aspect of human-computer interaction \cite{brave2007emotion}, especially with large language models (LLMs)~\cite{ratican2023six, sabour2024emobench}, as generating suitable emotional responses is essential for making communications natural and machines more human-like \cite{hortensius2018perception,li2022c3kg,kang2024nadine,sun2024fostering}. Expressing human-like emotions is challenging for machines, as it involves a complex psychological process that requires considering personal traits, situational influences, and an individual’s evaluation of the current scenario\cite{masters1991strategies,greenaway2018context,winberg2014stimulating}. Furthermore, emotion does not exist as a single entity but is generated as a collection of subjective experience and behavior \cite{gross2011emotion}.

Existing LLMs may fall short of capturing the nuances of human emotions, making their interactions seem impersonal or inadequate. For instance, in Figure \ref{fig:comparison}, we present an example conversation that showcased each speaker’s background and utterance. As we can see, GPT-4 \cite{openai2024gpt4technicalreport} and Llama3-8B-Chinese \cite{shenzhi_wang_2024} might generate responses that, although contextually relevant, fail to align with the emotional tone required in the given setting. This inadequacy stems from a lack of depth in understanding and simulating the complex process of human emotion generation. Furthermore, existing studies have primarily concentrated on English, despite the pivotal role of language in conveying emotions \cite{lindquist2017role}. Research on emotional expression in other languages and cultures has received limited attention in comparison.

To address this gap in emotional understanding, we turn to Cognitive Appraisal Theory (CAT) \cite{lazarus1991emotion}, a psychological framework that comprehensively explains how emotions are generated through the appraisal of external stimuli. At the core of CAT is the concept of \textit{appraisal}, which describes how individuals evaluate external stimuli, thereby eliciting emotional responses. This process underscores the dynamic role of interpretation in shaping emotions—such as when encountering an animal that a person believes may hurt him or her, the person might immediately appraise it as a threat, triggering fear. 

Building on CAT, we introduce \textbf{\PipeName{}}: a \textbf{C}ognitive \textbf{A}ppraisal \textbf{T}heory \textbf{Ba}sed \textbf{E}motionally \textbf{A}ppropriate \textbf{R}esponse framework designed to enhance LLMs' ability to express emotions accurately. \PipeName{} comprises three main components: \textit{1) intra-individual factors}, including personal factors and situational construal; \textit{2) the appraisal process}, which involves evaluating external stimuli based on these factors and the conversation history; and \textit{3) appraisal outcomes}, which are the resulting emotions and action tendencies. This structured approach ensures broad applicability across diverse populations and contexts, integrating emotions and actions into the ongoing conversation history. Given the importance of cultural context in emotional expression, our study specifically focuses on the use of Chinese, allowing us to tailor the dialogues to align with the nuances of Chinese language and culture.

Using the \PipeName{} framework, we generate a dataset of multi-turn dialogues between two individuals, each assigned a unique personality, goal, and situational construal. GPT-4-turbo generates relevant beliefs and knowledge to enrich each person's profile. Using this background, it sequentially produces individual emotion labels and utterances according to appraisal process guidelines. This automatic data synthetic framework yields \textbf{\DatasetName{}} (\textbf{C}ognitive \textbf{Ap}praisal theory-based \textbf{E}motional corpus), a dataset of 2,848 multi-turn dialogues covering 15 distinct emotions. The raw dataset undergoes a thorough cleaning and rigorous human evaluation. We utilize evaluation metrics from related studies—label correctness, emotion-utterance alignment, emotion-context alignment, intensity, coherence, and fluency—and enlist three raters to assess data quality. This process ensures accuracy in emotion labeling, contextual coherence, and overall conversational fluency.

%We use \PipeName{} to generate raw dialog data, followed by thorough data cleaning and careful manual evaluation. The resulting dataset, \textbf{\DatasetName{}} (\textbf{C}ognitive \textbf{Ap}praisal theory based \textbf{E}motional corpus), includes 2,848 dialogues. To assess the effectiveness of \DatasetName{} and \PipeName{}, we conduct ablation studies and utilize the dataset to develop \textbf{\ModelName{}}, a model specifically designed to engage in unseen conversations while expressing human-like emotions. We carry out both automatic and human evaluations to determine if \ModelName{}’s generated emotions and responses aligned more closely with ground truth compared to other baselines.

%A challenge is that emotions aren't always strictly separate categories; rather, they exist on a spectrum where some emotions are more similar to each other than others. Traditional evaluation metrics like exact-match treat emotions as entirely separate, discrete entities, failing to capture these subtleties. For example, if the true emotional response is "happiness," a generated response of "interest" might be seen as more similar or closer to "happiness" than "sadness" is. However, general metrics would not recognize this graded similarity and instead would categorize all incorrect answers as equally wrong. 

The practical use of dataset \DatasetName{} is shown through the creation of a fine-tuned model that can engage in new dialogues while effectively expressing human-like emotions. We test the model's performance through two tasks: predicting the next speaker's emotion label and their utterance in a dialogue.
For the first task, we evaluate how well the predicted emotion labels match the actual labels. Since traditional metrics, like exact-match, treat emotions as separate categories and ignore their nuanced nature, we include emotional distance as an additional metric to capture subtle similarities more effectively.
In the second task, we compare how closely the utterances generated by our model align with the ground truth. Through these two tasks, we show that our dataset enables the model to effectively capture and convey nuanced human emotions in dialogues. In summary, our study offers the following contributions:
\begin{itemize}[leftmargin=*]
    \item We propose \PipeName{}, an automatic data generation framework based on cognitive appraisal theory that addresses the challenge of aligning generated emotions with human-like emotions in dialogues.
    \item We then construct an emotional dialogue dataset, \DatasetName{}, with rigorous quality control and human evaluation to ensure accurate emotion labels and contextually appropriate utterances.
    \item We design two evaluation tasks: predicting the next speaker's emotion label and generating an appropriate emotional response. Our model fine-tuned on \DatasetName{} significantly outperforms state-of-the-art models in both tasks. We believe our framework advances research on CAT-theory-driven dialog systems for more human-like emotional responses.
\end{itemize}

\section{Related works}
Most related studies focus on three topics: textual emotion recognition in conversation, emotional support conversation, and emotion generation.
\subsection{Textual Emotion Recognition}

Textual emotion recognition (TER) aims to automatically identify emotions in text and has become a significant focus in natural language processing due to its academic and commercial potential. TER requires in-depth analysis and ongoing optimization of methodologies for accurate emotion prediction \cite{deng2021survey}. Moreover, the complexity of this topic grows from sentence level to document level, as emotions may be conveyed through subtle meanings, metaphors, sarcasm, and irony \cite{alswaidan2020survey}. Researchers have applied LLMs to recognize emotions from online posts \cite{liu2024emollms,yang2023towards}, TV series \cite{zhang2024refashioning,peng2024customising}, daily conversations \cite{fu2024ckerc}, and domain-specific dialogues \cite{xing2024designing}. To improve the accuracy of emotion recognition, many studies have begun to utilize multi-modal inputs, such as text, audio, and video \cite{zhang2023dialoguellm,lei2023instructerc,gan2023ziya2,cheng2024emotion}. Despite these advances, there is still room to explore how LLMs can authentically convey emotions for more human-like expressions. Our work addresses this challenge by using cognitive appraisal theory for emotion generation, providing a novel approach to enhancing human-LLM interactions.

\subsection{Emotional Support Conversation}
Emotional Support Conversation (ESC) aims to offer support to individuals dealing with emotional concerns through social interactions, emphasizing both conversational skills and counseling strategies \cite{liu2021towards}.
Researchers have employed LLMs to alleviate emotional problems \cite{zheng2023building} or elicit positive emotions \cite{zhou2023facilitating}. Recent studies highlight the importance of individual factors in formulating empathetic responses that better align with people's needs. Researchers took help-seekers’ persona \cite{cheng2022pal} and situational information \cite{sabour2022cem} into consideration to offer help across different populations. Furthermore, there have been studies analyzing users’ cognition and affection \cite{zhou2022case} or exploring emotion’s cause \cite{yang2024enhancing} to provide more appropriate empathetic responses. Current research on ESC emphasizes formulating empathetic responses but focuses mainly on understanding users' emotional states rather than enabling LLMs to express human-like emotions.

\subsection{Emotional Utterance Generation}
Studies on emotional utterance generation aim at aligning general human affective responses. Some promising studies adapted related psychological theories to let LLMs generate emotions. For example, Li et al. \cite{li2024enhancing} proposed ECoT that enhances LLMs’ emotional intelligence by incorporating Goleman’s Emotional Intelligence Theory~\cite{goleman2020emotional}. Moreover, \citet{croissant2024appraisal} considered both appraisal and memory systems to simulate affective outputs. While these studies demonstrate the effectiveness of their methods, they face limitations when adapted to real-life conditions across diverse populations: either intra-individual factors are not considered, or the context is confined to gaming scenarios.

\section{Method}
\subsection{Definitions}
\label{method:framework}
\begin{figure*}[!ht]
\scalebox{0.95}{
\includegraphics[width=\linewidth]{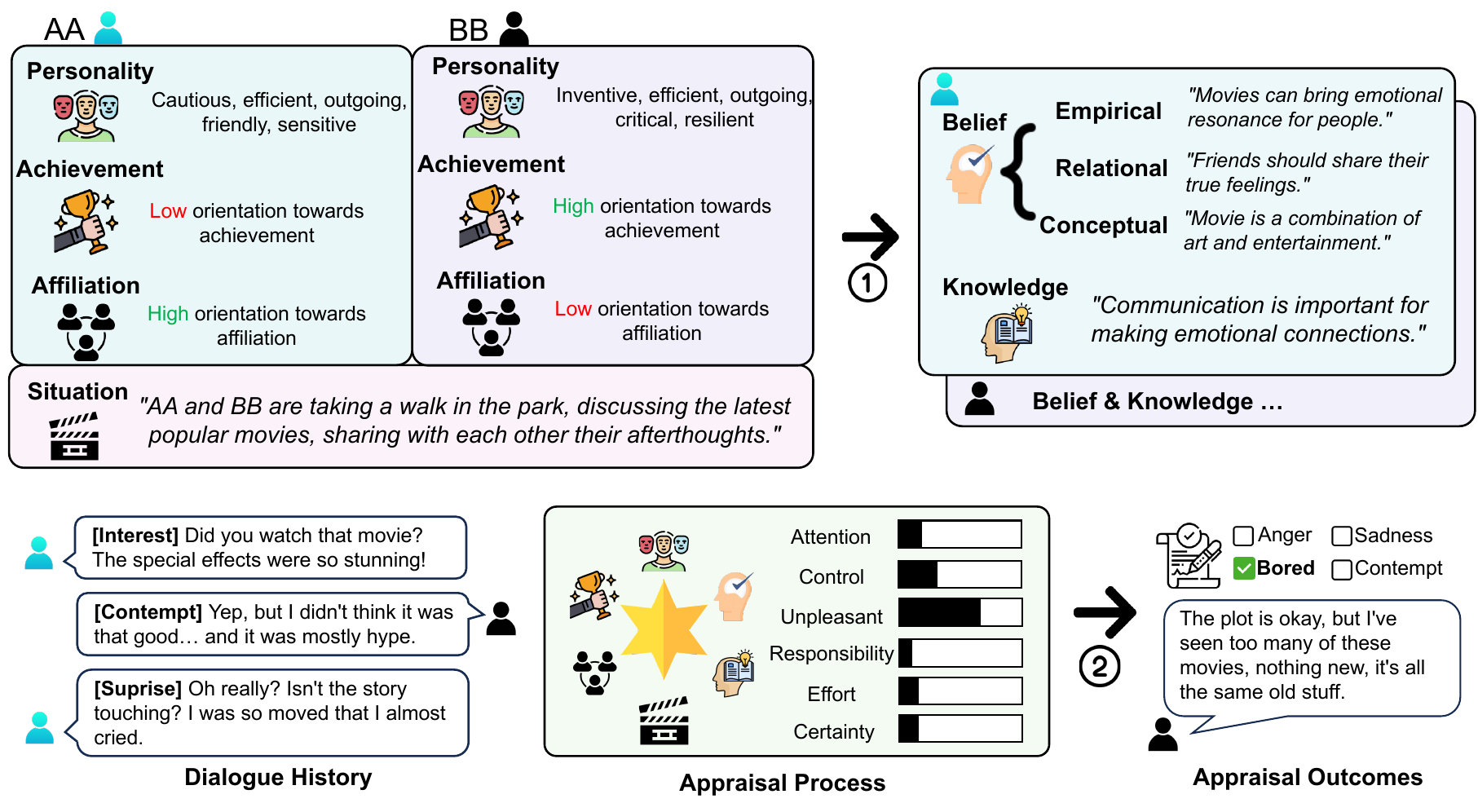}}
  \vskip -0.1in
  \caption{Overview of \PipeName{}. It contains two stages: (1) AA and BB are initially assigned specific personalities, goals, and situational construal, which are used to generate their beliefs and knowledge (2) the appraisal process, where individuals evaluate the interaction across six dimensions (unpleasantness, control, responsibility, certainty, effort, and attention) to generate emotions and utterances. }
  \label{fig:pipeline}
  \vskip -0.1in
\end{figure*}

In Cognitive Appraisal Theory, intra-individual factors shape appraisal outcomes through the appraisal process. These intra-individual factors include situational construal and personal characteristics such as personality, goals, beliefs, and knowledge. Appraisal outcomes, in turn, represent the results generated from this appraisal process. Below, we define each of these concepts in detail:
\begin{itemize}[leftmargin=*]
\setlength\itemsep{0.2em}
\item \textbf{Situational Construal} refers to individuals' representation of situations or their perception of situational variables \cite{funder2016taking}. Unlike the objective situation, situational construal is more subjective and interacts directly with personal factors. To cover as many conditions as possible, we use the 89 different situational construals from the Riverside Situational Q-sort \cite{funder2016taking}. The completed list of all 89 situational construals can be found in Appendix \ref{appendix:situation}.

\item \textbf{Personal Factors}. (1) \textit{Personality} are used from the 32 personalities defined in the Big Five personality traits \cite{roccas2002big}. (2) \textit{Goals} are orientations toward specific pursuits and fall into two main types: achievement goals and affiliation goals. Achievement goals focus on success and the effort needed for performance \cite{elliott1988goals}, while affiliation goals drive the desire for social contact, prompting individuals to seek emotional support, attention, and positive reinforcement \cite{hill1987affiliation}. (3) \textit{Beliefs} are what is normatively acceptable, feasible, and legitimate and can lead to various emotions \cite{lazarus1991emotion}. It can be divided into empirical (to believe objects or perceived value), relational (to believe someone), and conceptual (to believe in narratives) beliefs \cite{seitz2020belief}. (4) \textit{Knowledge} refers to the understanding of potential harms or benefits \cite{lazarus1991emotion}. It can significantly influence how an encounter is appraised as beneficial or not, thereby leading to different emotions. 

\item \textbf{Appraisal Outcomes} includes both emotion and corresponding behavior (i.e., utterances in the dialogue). We employ the 15 emotions from Smith and Ellsworth's study \cite{smith1985patterns}: happiness, sadness, anger, boredom, challenge, hope, fear, interest, contempt, disgust, frustration, surprise, pride, shame, and guilt. 
\end{itemize}

\subsection{Data-generative Framework}

In this part, we present the data-generative framework \PipeName{} for simulating dialogue between two individuals. As depicted in Figure \ref{fig:pipeline}, the framework unfolds in two stages. 

\noindent
\textbf{First Stage: generating beliefs and knowledge based on Personality, Goals, and Situation.} In this stage, each speaker is assigned a unique personality and set of goals. For example, AA is defined as "cautious, efficient, outgoing, friendly, and sensitive", with low achievement and high affiliation goals. Conversely, BB is "inventive, critical, and resilient", with high achievement and low affiliation goals. The pre-defined situational context—such as {"a casual walk in the park and a discussion about popular movies"}—is then enriched with these personality traits and goals, providing a more detailed background for the dialogue. Using these attributes, we further prompt GPT-4-turbo-0409 to generate personalized beliefs and knowledge for each individual, as belief and knowledge generation are influenced by an individual's unique characteristics in specific contexts \cite{lazarus1991emotion}. The detailed prompt of belief and knowledge generation is listed in Appendix \ref{appendix:belief_knowledge}. For instance, AA's belief that “friends should share their true feelings” aligns with a high affiliation goal and the context of reflecting on shared experiences.

\noindent
\textbf{Second Stage: simulating appraisal process to generate emotion label and utterance.} In this stage, we model the appraisal process, where each speaker assesses their experience based on six dimensions: unpleasantness, control, responsibility, certainty, effort, and attention. 
By analyzing these dimensions alongside the dialogue history and predefined intra-individual factors, we design a targeted prompt for GPT-4-turbo to predict the speaker's emotion. The model sequentially categorizes the degree of each dimension—beginning with unpleasantness, followed by expected effort, attention, uncertainty, control, and responsibility—to identify the most probable emotion label and the corresponding emotion-aligned utterance.
Detailed prompts describing this appraisal process are provided in the Appendix \ref{appendix:emotion_guideline}. For example, as illustrated in the bottom right of Figure~\ref{fig:pipeline}, BB's appraisal, marked by high unpleasantness and low scores on other dimensions, suggests an emotion of \textit{"boredom"} and an utterance expressing disinterest in the movie, aligning with BB's personal factors, situational understanding, and dialogue history. 
% This method ensures that the generated emotions are accurate, appropriate, and contextually coherent.

\subsection{Data Quality Control}
\vskip -0.1cm
\noindent
\textbf{Manual Dialogue Refinement.} To improve data quality, we recruited three native Chinese-speaking workers, all of whom are university students or graduates, to meticulously review and refine dialogues generated by LLMs. Each worker carefully read the instructions and completed trial annotations before beginning the task. The manual filtering process involved: (1) cleaning emotion labels by removing irrelevant ones and aligning labels with the settings, and (2) refining utterances to match emotion labels and modifying phrasing for colloquial language and authentic Chinese conversational tone. For cost reasons, each dialogue was assigned to a single worker, and we randomly selected subsets for quality checks, comparing raw and refined versions.

\noindent
\textbf{Data Quality Evaluation.} We recruited six additional Chinese-speaking workers, split into two groups, to assess raw and refined data quality. The evaluation metrics, based on prior research \cite{qian2023think} and tailored to our study, include:
(1) \textit{EmoCategory}, assessing whether the emotion label aligns with the settings; (2) \textit{EmoMatch}, evaluating the degree to which the utterance conveys the labeled emotion; (3) \textit{SettingMatch}, measuring if the utterance content aligns with intra-individual factors and situational context; (4) \textit{EmoIntensity}, indicating the intensity of the labeled emotion in the utterance; (5) \textit{Coherence}, determining if the utterance logically fits the conversation context; and (6) \textit{Fluency}, assessing whether the utterance is fluent and easy to understand.
 
We further curated a human evaluation set by randomly selecting two utterances from each dialogue in the test set, resulting in a total of 283 utterances for rating. Each utterance is provided with a corresponding emotion label, intra-individual factors, and situational construal. We provide each rater with a detailed rating manual outlined in the Appendix \ref{appendix:dataset_human_evaluation}. The quality of the human evaluation for the dataset, before and after refinement, is summarized in Table~\ref{tab:human_eval_on_cape}. Correlation between raters' scores is significant across dimensions ($p < 0.05$). We observe that manual calibration improves the accuracy of emotion labels and ensures that utterances better align with character settings and dialogue context.

%The results of the human evaluation (from Table \ref{tab:human_eval_on_cape}) show that in the test set, 88. 89\% of the emotion labels generated by ChatEMO fit the best among all emotion choices for the responder, reflecting its effectiveness in predicting reasonable emotion.
\begin{table}
  \centering
  \small
  \setlength{\belowcaptionskip}{-0.3cm}
  \scalebox{0.95}{
  \begin{tabular}{lccc}
    \toprule
    {Dimensions} & {\makecell{Score\\Range}} & {\makecell{Ratings \\before Filtering}} & {\makecell{Ratings \\after Filtering}}\\
    \midrule
    EmoCategory & 0-1 & 0.89 & 0.93 (\textcolor{teal}{$\uparrow$4.5\%})\\
    EmoMatch & 1-5 & 4.61 & 4.80 (\textcolor{teal}{$\uparrow$4.0\%})\\
    SettingMatch & 1-5 & 4.09 & 4.52 (\textcolor{teal}{$\uparrow$9.5\%})\\
    EmoIntensity & 0-2 & 1.76 & 1.80 (\textcolor{teal}{$\uparrow$2.3\%})\\
    Coherence & 1-5 & 4.93 & 4.94\\
    Fluency & 1-5 & 4.80 & 4.85 (\textcolor{teal}{$\uparrow$1.0\%})\\
    \bottomrule
  \end{tabular}}
  \caption{Utterance-level human evaluation on CAPE.}
  \label{tab:human_eval_on_cape}
\end{table}

\subsection{Dataset Statistics}

%We built a multi-turn conversational corpus named \textbf{\DatasetName{}} (\textbf{C}ognitive \textbf{Ap}praisal theory based \textbf{E}motional corpus) based on the \PipeName{}. For each of the 89 situational construals, we randomly chose personalities and goals for the two characters in the conversations and promised that all kinds of personalities and goals were covered. Beliefs and knowledge are then generated for each character based on the situational construals, one’s personality, and goals. 

%Then all these intra-individual factors and an appraisal guideline (mentioned in \ref{method:framework} and Appendix \ref{appendix:emotion_guideline}) are provided to GPT-4 to choose the emotion of the first character and generate action (that is, the character's utterance). The emotion and utterance will be added to the conversation history. GPT-4 will then read the conversation history and choose the other character’s emotion, generate and add this character’s utterance to the conversation history. This process will be repeated until there are around five rounds. Finally, GPT-4 polishes all the conversations following a colloquial guideline that we designed. 

\begin{figure}[!ht]
\centering
\setlength{\belowcaptionskip}{-0.3cm}
\includegraphics[width=1.0\linewidth]{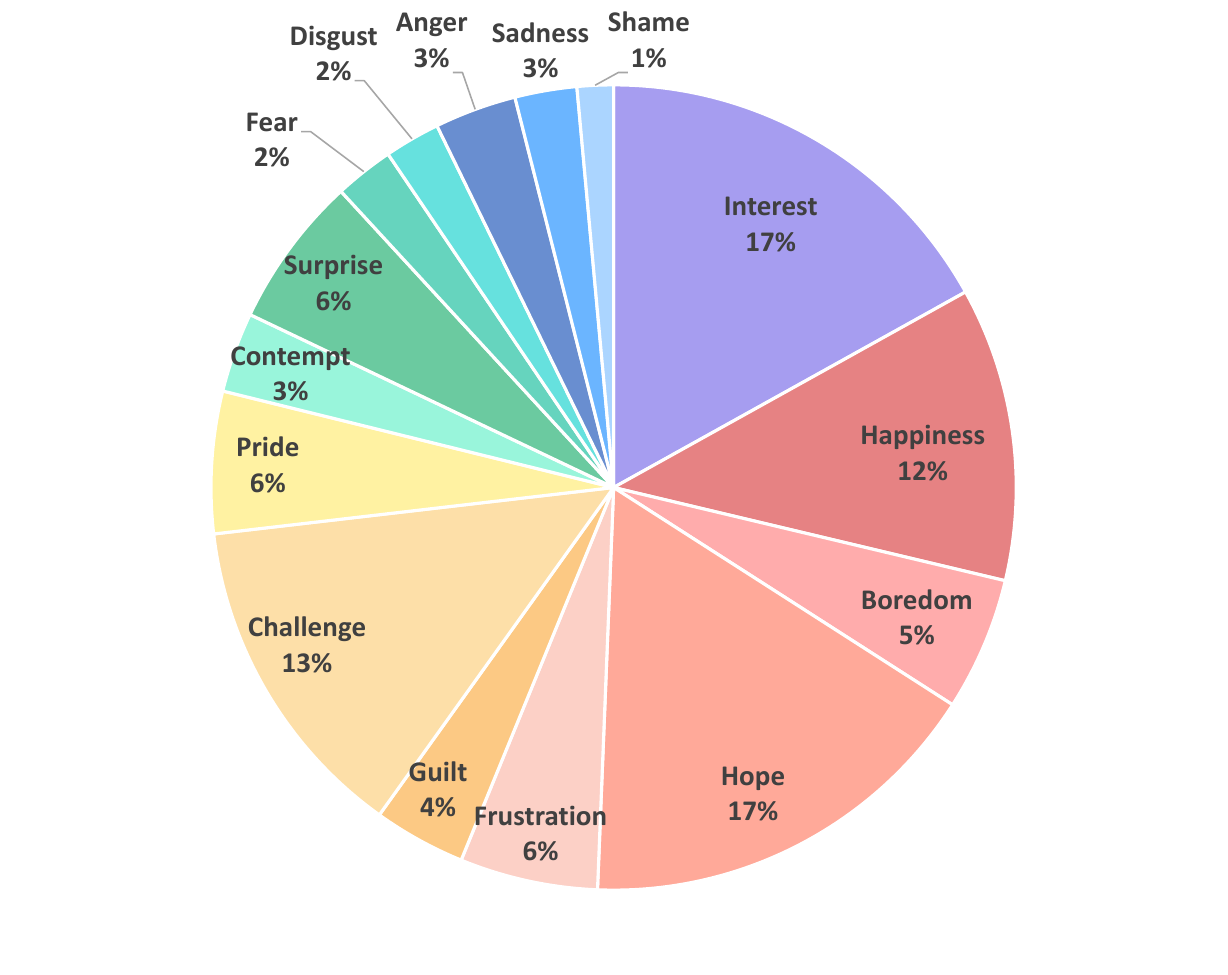}
  \vskip -0.2in
  \caption{Emotions distribution of \DatasetName{}}
  \label{fig:distribution}
\end{figure}

We named the refined dataset as \DatasetName{} (Cognitive Appraisal theory-based Emotional corpus). Table \ref{tab:statistics} offers a comprehensive overview of the \DatasetName{} dataset, encompassing 2,848 dialogues which cover 89 unique situations. The distribution of emotions of \DatasetName{} can be found in Figure~\ref{fig:distribution}. The average number of utterances per dialogue is around 10, with an average of 38.3 tokens per utterance, indicating a substantial amount of information exchanged within the conversations. 

% \begin{table}[!ht]
% \centering
% \small
% \scalebox{1.0}{
% \begin{tabular}{lcccc}
%     \toprule
%     {Aspects} & {{Train}} & {Test} & {Validation} & {Total}\\
%     \midrule
%     \# Dialogues & 2,563 & 142 & 143 & 2,848 \\
%     Avg. \# turns per dialog & 10 & 10 & 10 & 10 \\
%     Avg. \# tokens per utterance & 21.2 & 21.6 & 20.9 & 21.3 \\
%     \bottomrule
% \end{tabular}
% }
% \caption{
% \DatasetName{} dataset statistics.
% }
% \label{tab:statistics}
% \end{table}

\begin{table}[!ht]
\centering
\setlength{\tabcolsep}{17pt}
\setlength{\belowcaptionskip}{-0.3cm}
\small
\begin{tabular}{l|c}
\toprule
\# Dialogs                           & 2,848  \\
\# Utterances                        & 28,643 \\
\# Situations                        & 89     \\ \midrule
Avg. \# utterances per dialog        & 10.0   \\
Avg. \# tokens per dialog            & 385.0  \\
Avg. \# tokens per utterance         & 38.3   \\ \midrule
\# Dialogs in Train Set  & 2,563  \\
\# Dialogs in Validation Set  & 143  \\
\# Dialogs in Test Set  & 142  \\ \bottomrule
\end{tabular}%
\vskip -0.2cm
\caption{Statistics of the \DatasetName{} dataset.}
\label{tab:statistics}
\end{table}

To evaluate the quality of \DatasetName{}, we compare it with representative emotional dialogue datasets, as shown in Appendix \ref{appendix:dataset_comparison}. The results indicate that \DatasetName{} is uniquely grounded in psychological theories and incorporates a broad range of personal factors closely linked to emotions. It also boasts wider coverage of emotions and situations compared to existing datasets. Furthermore, unlike datasets annotated from TV series dialogues, \DatasetName{} is generated synthetically, which frees it from specific character constraints and limited kinds of story lines, allowing for a more flexible and diverse approach to dialogue generation.

\section{Experiments}
\subsection{Tasks}
We have outlined two tasks to evaluate the performance of our dataset, including emotion prediction and appropriate emotional utterances generation. 

\noindent
\textbf{Task1: Emotion Prediction.} This task is to let LLMs to predict the emotion label of the next speaker in the dialogue based on intra-individual factors and dialogue history. The predicted emotion label should be chosen within the 15 emotions defined in the \PipeName{}. It requires LLMs to understand and anticipate the emotions a person may express under specific conditions.

\noindent
\textbf{Task2: Next Utterance Prediction.} The task asks LLMs to generate the utterance of the next speaker in the dialogue based on intra-individual factors and dialogue history. The quality of the generated utterances reflects a model's ability to express human-like emotional responses. 
% Given the subjective nature of utterance evaluation, we try to enhance our evaluation by randomly selecting 200 utterances from top-performing baselines and recruiting workers to evaluate them. We incorporate the dimensions including SettingMatch, Coherence, and Fluency from the dialogue quality evaluation section to guide the human rating process.

\subsection{Models}
\noindent
\textbf{Models for Task1.}
When evaluating the capability of predicting emotions, we compared our model with baselines with high proficiency in Chinese including GLM-4-0520 \cite{glm2024chatglm}, GLM-4-9B-chat~\cite{glm2024chatglm}, Llama3-8B-Chinese-Chat~\cite{shenzhi_wang_2024}, DeepSeek-v2~\cite{deepseekv2}, Qwen-2.5-72B-Instruct~\cite{qwen2.5}, and GPT-4-0613~\cite{openai2024gpt4technicalreport}. For all LLMs, we report zero-shot and 4-shot results.  

\noindent
\textbf{Models for Task2.} 
When examining the generation of appropriate emotional utterances, we incorporated two expert models alongside the LLMs mentioned in the emotion prediction task. The first is CharacterGLM-3 \cite{CharacterGLM} which supports role-playing based on character settings, and the second one is Emohaa \cite{Emohaa} which is designed to provide empathetic support. Following the specified input requirements for inference, we report only zero-shot results for both CharacterGLM-3 and Emohaa. For the rest LLMs, we report both zero-shot and 4-shot results.

\noindent
\textbf{\ModelName{}.} We further conduct supervised fine-tuning of the GLM-4-9B-Chat model on both two tasks to develop our model. We conducted experiments on 8 Nividia RTX A6000 GPUs, setting the total batch size as 64, with gradient accumulation as 4, the maximum learning rate as 1e-5, and the total number of training epochs as 3.

\begin{table*}[!ht]
\small
  \centering
  \setlength{\tabcolsep}{13pt}
  \scalebox{0.92}{
  \begin{tabular}{l|c|ccccc}
    \toprule
    {{Models}} & {{Setting}} & {{Acc. $\uparrow$}} & {{F1} $\uparrow$} & {{Precision}$ \uparrow$} & {{Recall} $\uparrow$} & {{CAT-Dist} $\downarrow$} \\
    \midrule
     GPT-4-0613 & \multirow{5}{*}{\shortstack{zero-shot}} & 0.21 & 0.14 & 0.18 & 0.14 & 0.218 \\
    GLM-4-0520& & 0.18 & 0.12 & 0.24 & 0.13 & 0.229 \\
    DeepSeek V2 & &0.22 & 0.18 & 0.21 & 0.19 & 0.221 \\
    Qwen-2.5-72B & &0.17 & 0.13 & 0.14 & 0.13 & 0.247 \\
    Llama3-8B-Chinese & &0.12 & 0.07 & 0.08 & 0.11 & 0.265 \\
    GLM-4-9B & &0.12 &0.09  &0.09 &0.11 &0.268 \\
    \midrule
     GPT-4-0613 & \multirow{5}{*}{\shortstack{4-shot\\prompting}} & 0.25 & 0.16 & 0.20 & 0.16 & 0.213 \\
     GLM-4-0520& &\underline{0.27} & 0.19 & 0.26 & 0.19 & \underline{0.207} \\
     DeepSeek V2 & &0.26 & \underline{0.20} & \textbf{0.30} & \underline{0.21} & 0.213 \\
     Qwen-2.5-72B & &0.24 & 0.19 & 0.20 & 0.20 & 0.224 \\
     Llama3-8B-Chinese & &0.14 & 0.06 & 0.10 & 0.08 & 0.268 \\
     GLM-4-9B & &0.17 &0.10  &0.11 &0.11 &0.242 \\
    \midrule
    \textbf{\ModelName} & fine-tuning &\textbf{0.36} & \textbf{0.28} & \underline{0.29} & \textbf{0.28} & \textbf{0.181} \\
    \bottomrule
  \end{tabular}}
\vskip -0.1in
  \caption{
    Performance comparison on Emotion Prediction.
{(\small \textbf{Bold} for best, \underline{underline} for the second best).}
  }
  \label{tab:task1}
\end{table*}

% \begin{table*}[!ht]
% \small
%   \centering
%   \begin{tabular}{clccccc}
%     \toprule
%     \textbf{Setting} & \textbf{Model} & \textbf{Accuracy} & \textbf{F1} & \textbf{Precision} & \textbf{Recall} & \textbf{DistanceScore} \\
%     \midrule
%     \multirow{5}{*}{zero-shot} & GPT-4 & 0.21 & 0.19 & 0.21 & 0.21 &  \\
%     & GLM-4 & 0.18 & 0.15 & 0.20 & 0.18 &  \\
%     & DeepSeek & 0.17 & 0.16 & 0.17 & 0.17 &  \\
%     & Qwen-2.5 & 0.22 & 0.21 & 0.24 & 0.22 &  \\
%     & Llama3-Chinese & 0.12 & 0.10 & 0.11 & 0.12 &  \\
%     \midrule
%     \multirow{5}{*}{few-shot} & GPT-4 & 0.25 & 0.24 & 0.27 & 0.25 &  \\
%      &GLM-4 & 0.27 & 0.24 & 0.26 & 0.27 &  \\
%      &DeepSeek & 0.24 & 0.23 & 0.24 & 0.24 &  \\
%      &Qwen-2.5 & 0.26 & 0.24 & 0.31 & 0.26 & \\
%      &Llama3-Chinese &  &  &  &  &  \\
%     \midrule
%     & \textbf{\ModelName}  &\textbf{0.37} & \textbf{0.35} & \textbf{0.37} & \textbf{0.36} & \textbf{} \\
%     \bottomrule
%   \end{tabular}
%   \caption{\label{citation-guide}
%     Performance comparison between ChatEMO and baselines on predicting emotions.
%   }
% \end{table*}
\begin{table*}[!ht]
  \centering
  \small
  \scalebox{0.88}{
  \begin{tabular}{l|c|cccccc}
    \toprule
    {Models} & {Setting} & {BLEU-1} $\uparrow$ & {BLEU-2} $\uparrow$ & {ROUGE-1} $\uparrow$ & {ROUGE-2 $\uparrow$} & {ROUGE-L} $\uparrow$ & {BERTScore} $\uparrow$ \\
    \midrule
    GPT-4-0613 & \multirow{7}{*}{\shortstack{zero-shot}} &23.63 & 5.13 & 22.71 & 2.66 & 19.10 & 61.07 \\
    GLM-4-0520 & &22.46 & 4.76 & 24.24 & 3.66 & 19.76 & 61.61 \\
    DeepSeek V2 & &24.60 &5.38 & 24.49 & 3.45 & 20.11 & 61.36 \\
    Qwen-2.5-72B  & &24.45 &5.36 & 23.48 & 3.46 & 19.46 & 61.55 \\
    Llama3-8B-Chinese & &14.57 &3.30 & 17.20 & 1.55 & 13.45 & 57.23 \\
    Emohaa & &17.91 &3.88 & 22.72 & 2.68 & 17.43 & 60.27 \\
    CharacterGLM3 & &25.67 &\textbf{5.84} & 23.89 & 3.53 & 19.94 & 61.43 \\
    GLM-4-9B  & &24.06 &5.55 &21.49  &2.20 &17.90 &60.39 \\
    \midrule
    GPT-4-0613 & \multirow{5}{*}{\shortstack{4-shot\\prompting}}&\textbf{27.16} &\underline{5.74} & 25.52 & 3.68 & 21.39 & 62.40 \\
    GLM-4-0520 & &25.11 &5.42 & 24.63 & 3.52 & 20.46 & 61.66 \\
    DeepSeek V2 & &25.83 &5.47 & \textbf{26.40} & 4.22 & \underline{21.86} & 62.26 \\
    Qwen-2.5-72B & &26.02 &5.53 & 25.92 & \underline{4.37} & 21.78 & \underline{62.41} \\
    Llama3-8B-Chinese & &14.29 &3.26 & 17.08 & 1.34 & 13.67 & 56.59 \\
    GLM-4-9B  & &23.43 &5.44  &22.19 &2.20 &18.11 &60.78  \\
    \midrule
    \textbf{\ModelName} & fine-tuning &\underline{26.46} &5.59 & \underline{26.36} & \textbf{4.90} & \textbf{22.35} & \textbf{62.83} \\
    \bottomrule
  \end{tabular}
  }
  \vskip -0.1in
  \caption{
    Performance comparison on generating emotional utterance. {(\small \textbf{Bold} for best, \underline{underline} for the second best).}
  }
  \label{tab:task2}
\end{table*}

\subsection{Evaluation Metrics}
\noindent
\textbf{Metric for Task1.} According to previous studies \cite{chen2022cped,wen2021automatically,shen2020memor}, we used general metrics including Accuracy, Macro F1, Precision, and Recall to measure LLMs' performance on the emotion prediction task. Moreover, to address the limitation of general metrics that treat emotions discretely and overlook their nuanced spectrum-like nature, we introduced \MetricsName{} as an additional metric to capture subtle similarities more effectively. \MetricsName{} is inspired by the study by Smith and Ellsworth \cite{smith1985patterns}, which mapped 15 emotions on the six dimensions (of the appraisal process). Each emotion is represented as a six-dimensional vector, and the specific values can be found in detail in Appendix \ref{appendix:emotion_score}. We define \MetricsName{} as the average Manhattan distance between two emotion vectors across all six dimensions:
\vskip -0.5cm
\begin{equation*}
\operatorname{\MetricsName{}}(E_1, E_2) = \frac{1}{D} \sum_{i=1}^{D} \left| E_{1,i} - E_{2,i} \right|
\end{equation*}
where $E_1=(E_{1,1}, E_{1,2}, \dots, E_{1,D})$ and $(E_2 = (E_{2,1}, E_{2,2}, \dots, E_{2,D})$ 
 are the two emotion vectors, $D$ is the number of dimensions (here, \( D = 6 \)). 
% offers interpretability by highlighting dimension-specific differences, captures nuanced emotional variations across all six dimensions, and remains computationally efficient.

\noindent
\textbf{Metric for Task2.}
Following previous studies \cite{varshney2021modelling,chen2022cped}, we choose
the metrics below: (1) BLEU-1/2 \cite{papineni2002bleu} that evaluates the model's output quality by comparing it to references through 1/2-gram matching; (2) ROUGE-1/2/L \cite{lin2004rouge} that measures the longest common subsequence between output and reference, and (3) BERTScore \cite{zhang2019bertscore} which assesses output quality by cosine similarity of BERT embeddings. 

% To further enhance our evaluation, we randomly select 200 utterances generated by top-performing baselines based on general metrics and recruit workers to rate both their and \ModelName{}'s outputs. Given that task2 does not have emotion labels, we adapt dimensions including SettingMatch, Coherence, and Fluency from the dialogue quality evaluation section to guide the human rating process.

\section{Results}
\subsection{Emotion Prediction}
%Table~\ref{tab:task1} highlights \ModelName{}'s superior performance across most metrics. In baselines, GLM-4-0520 and DeepSeekV2 demonstrate competitive performance, especially when using few-shot learning techniques. Our results show that open-source models generally surpass their closed-source counterparts in terms of prediction accuracy. Additionally, larger models with more parameters tend to achieve higher accuracy and F1 scores compared to their smaller counterparts. We also observe a difference between Chinese-based LLMs and those not specifically tailored for Chinese data. The Chinese-based models exhibit a more significant improvement with few-shot examples. This suggests a heightened sensitivity of these models to Chinese contextual examples during inference.

The results of the emotion prediction task are shown in Table \ref{tab:task1}. Larger LLMs (i.e., GPT-4-0613, GLM-4-0520, DeepSeek V2, and Q-2.5-72B) perform better than smaller LLMs (i.e., Llama3-8B-Chinese and GLM-4-9B), due to more parameters and extensive pre-training data. Among large models, DeepSeek V2 achieves the highest accuracy and F1 score in zero-shot settings, while GLM-4-0520 shows the highest performance boost in few-shot settings, with a remarkable 50\% increase in accuracy and 58.3\% increase in F1 score. \ModelName{} achieves the highest accuracy and F1 score, surpassing all baselines in zero-shot and few-shot settings. Supervised fine-tuning yields a twofold increase in both accuracy and F1 score compared to the original GLM-4-9B, demonstrating that our dataset provides comprehensive information necessary for precise emotion prediction.
%Furthermore, \MetricsName{}, generally aligns well with exact-match metrics, though minor variances are present. In the few-shot setting, Qwen-2.5-72B outperforms DeepSeek V2 across all exact-match metrics. However, interestingly, Qwen-2.5-72B's performance on \MetricsName{} is lower than that of DeepSeek V2. This result indicates that \MetricsName{} may capture additional information beyond traditional exact-match metrics.

\subsection{Next Utterance Prediction}
Results of this task are shown in Table \ref{tab:task2}. Similar to the emotion prediction task, large LLMs continue to outperform small ones in predicting the next utterance. However, CharacterGLM-3, a 6B model, shows comparable performance to large models in both zero-shot and few-shot settings. This may be attributed to its specialized design for human-like role-playing. Conversely, Emohaa's suboptimal performance may stem from its emphasis on providing empathetic support, highlighting differences between giving empathetic responses and expressing emotional utterances. \ModelName{} performs the first or second-best across most metrics. Its proficiency lies in ROUGE-1/2/L metrics over BLEU-1/2, indicating precision in word identification but a tendency to generate irrelevant words. Notably, \ModelName{} ranks among the top performers in BERTScore, suggesting a high degree of similarity in content between its outputs and the ground truth.

%From the results in Table \ref{tab:task2}, \ModelName{} demonstrates either the best or second-best performance across most metrics. Its performance excels in ROUGE-1/2/L metrics compared to BLEU-1/2, indicating proficiency in identifying exact words but a tendency towards generating more unrelated words. Notably, \ModelName{} ranks among the top performers on BERTScore, suggesting the content of its outputs closely resembles the ground truth. 
\begin{table}
  \centering
  \small
  \setlength{\belowcaptionskip}{-0.3cm}
  \scalebox{0.95}{
  \begin{tabular}{lccc}
    \toprule
    {Models} & \makecell{SettingMatch\\(1-5)} & \makecell{Coherence\\(1-5)} & \makecell{Fluency\\(1-5)}\\
    \midrule
    GPT-4-0613 & \underline{4.52} & 4.51 & 3.85 \\
    CharacterGLM-3 & 4.11 & \underline{4.75} & \underline{4.03} \\
    DeepSeek V2 & 3.16 & 3.48 & 3.52 \\
    ChatEMO & \textbf{4.55} & \textbf{4.80} & \textbf{4.52}\\
    \bottomrule
  \end{tabular}}
  \vskip -0.1cm
  \caption{Human ratings on generated utterances.}
  \vskip -0.2cm
  \label{tab:task2human}
\end{table}

Given the subjective nature of utterance evaluation, the metrics above may not fully reflect the quality of generated responses. Therefore, we try to enhance our evaluation by randomly selecting 200 utterances from top-performing baselines and recruiting workers to evaluate them. \ModelName{}, GPT-4-013, CharacterGLM-3, and DeepSeek V2 were chosen as they achieve the best performance on at least one metric. We incorporate the dimensions including SettingMatch, Coherence, and Fluency from the dialogue quality evaluation section to guide the human rating process. The results can be found in Table \ref{tab:task2human}. To better demonstrate the comparison, we provide an example of generated utterances by the four LLMs in Appendix \ref{appendix:generated_utterance_comparison}.

%From Table~\ref{tab:task2}, the utterance provided by ChatEMO during the emotional conversations achieved the best performance on \text{ROUGE-2} (4.90), \text{ROUGE-L} (22.35) and \text{BERTScore} (62.83), though the \text{BLEU-2} (62.83) and \text{ROUGE-1} (26.36) are slightly lower than a few baselines. Since similar textual statements may reflect completely different emotions, human evaluation is a more reliable way to measure the appropriateness of the utterance. The high scores on EmoMatch (4.6111, max=5) and EmoIntensity (1.7593, max=2) show that ChatEMO can do both appropriate emotion generation while the intensity of the emotion is high enough to reflect the responder’s intra-individual factors.

\section{Analysis}
\subsection{Ablation: How Our Framework Design Enhances Dialogue Quality?}

We investigated how our data-generative framework enhances emotional dialogue quality. We omit key components of our framework to evaluate data quality without certain steps, comparing three versions: the full framework, one excluding belief and knowledge generation, and one without the appraisal process. Three Chinese-speaking workers are recruited to assess the quality of dialogues, based on five dimensions used in previous studies \cite{liu2024speak,zhang2024comprehensive,ou2023dialogbench,zhong2022towards}: (1) \textit{coherence}, how much the dialogue maintains logical consistency without confusion; (2) \textit{naturalness}, how well the dialogue aligns with Chinese speech habits; (3) \textit{correctness}, whether the label belongs to the given 15 emotions; (4) \textit{contextual relevance}, the degree to which the dialogue aligns with the settings; and (5) \textit{emotional relevance}, the alignment of emotional utterances with the settings. A rating guideline is provided for more reliable ratings, which is listed in the Appendix \ref{appendix:dialogue_rating_guideline}.

Human evaluation result (Table \ref{tab:pipe-ablation}) demonstrates that the absence of belief and knowledge highly impacts coherence and naturalness, as they contribute useful information to enrich the storyline and elevate the generated dialogues' quality. Moreover, omitting the appraisal process leads to notable decreases in correctness, contextual, and emotional relevance. This is because the appraisal process offers detailed guidelines for generating appropriate emotions and actions, ensuring that emotions align logically and dialogues fit the context.

%average kappa of the five dimensions is 0.2849
%all three raters' ratings on five dimensions are significantly correlated ($p < 0.05$)
\begin{table}[!ht]
  \centering
  \small
  \setlength{\tabcolsep}{3pt}
  \setlength{\belowcaptionskip}{-0.3cm}
  \scalebox{0.81}{
  \begin{tabular}{lcccc}
    \toprule
    {{Dimensions}} &{{\makecell{Score\\Range}}} & {\PipeName{}} & {\makecell{\text{w/o} Belief \&\\  Knowledge}} & {\makecell{\text{w/o}\\ Appraisal}} \\
    \midrule
    Coherence &1-5 & 4.81 & 4.29 (\textcolor{red}{$\downarrow$11\%}) & 4.32 (\textcolor{red}{$\downarrow$10\%}) \\
    Naturalness &1-5 & 4.66 & 4.02 (\textcolor{red}{$\downarrow$14\%}) & 4.14 (\textcolor{red}{$\downarrow$11\%}) \\
    Correctness &0-1 & 1.00 & 0.66 (\textcolor{red}{$\downarrow$34\%}) & 0.50 (\textcolor{red}{$\downarrow$50\%}) \\
    \makecell{Contextual Relevance} &1-5 & 4.90 & 4.28 (\textcolor{red}{$\downarrow$13\%}) & 4.18 (\textcolor{red}{$\downarrow$15\%}) \\
    Emotional Relevance &1-5 & 4.77 & 3.97 (\textcolor{red}{$\downarrow$17\%}) & 3.89 (\textcolor{red}{$\downarrow$18\%}) \\
    \bottomrule
  \end{tabular}
  }
  \vskip -0.2cm
  \caption{Human ratings on generated dialogues by different data synthetic frameworks.}
  \label{tab:pipe-ablation}
\end{table}

\subsection{How Emotion Labels Impact the Quality of Generated Utterances?}
To explore whether an agent can first self-predict emotion before generating a response—thereby enhancing emotional response generation—we draw on the emotion chain-of-thought method~\cite{li2024enhancing}. Our experiments consist of two parts: (1) \textit{Conditional Setting}, where we examine whether providing the ground-truth emotion label improves the quality of the generated utterance; and (2) \textit{Joint-Modeling Setting}, which encourages the model to first predict the emotion and then generate response. We perform supervised fine-tuning for both settings align as above. As shown in Table \ref{tab:ablation-task2}, when the emotion label is accurate, the quality of generated utterances significantly improves. Additionally, joint-modeling not only enhances emotion labeling accuracy but also results in contextually relevant and emotionally resonant responses. This implies that future work could leverage our dataset to develop a more sophisticated emotion CoT, further enhancing agent emotional intelligence.
\begin{table}[!ht]
  \centering
  \small
  \scalebox{0.88}{
  \begin{tabular}{lccc}
    \toprule
    {Models} & {Acc.} $\uparrow$ & {F1} $\uparrow$ & {CAT-Dist} $\downarrow$ \\
    \midrule
    \ModelName{} & 0.36 & 0.28  & 0.181 \\
    joint & 0.39(\textcolor{teal}{$\uparrow$8\%})  & 0.38(\textcolor{teal}{$\uparrow$36\%}) & 0.173(\textcolor{teal}{$\downarrow$4\%}) \\
    \midrule
    {Models} & {BLEU-1/2} $\uparrow$ & {ROUGE-1/2/L} $\uparrow$ & {BERTScore} $\uparrow$ \\
    \midrule
    \ModelName{} & 26.46 / 5.59 & 26.36 / 4.90 / 22.35 & 62.83 \\
    cond & \textbf{29.82} / \textbf{6.09} & \textbf{29.49} / \textbf{6.75} / \textbf{25.02} & 64.58 \\
    joint & 28.28 / 5.93 & 28.03 / 5.75 / 23.76 & \textbf{70.96} \\
    \bottomrule
  \end{tabular}
  }
  \vskip -0.1in
  \caption{
    Effects of emotion label prediction accuracy on response quality. (task1 above, task2 below).
  }
  \label{tab:ablation-task2}
\end{table}

\section{Conclusions}
In this paper, we introduced the Emotion Conversations Generation Framework, which harnesses the Cognitive Appraisal Theory to generate emotionally appropriate responses in conversational agents. Our framework delves into intra-individual factors such as situational construal and personality, incorporating a comprehensive appraisal process to enable nuanced emotion predictions and utterances generation. Building upon this framework, we curated a multi-turn conversational corpus, \DatasetName{}, comprising 2,848 emotional dialogues meticulously refined by human crowd workers. Utilizing \DatasetName{}, we fine-tuned a model and conducted experiments that showcased the superior performance of our fine-tuned LLM over various robust baseline models in both emotion prediction and utterance generation tasks. The results of our study underscore the transformative potential of integrating cognitive appraisal insights into LLMs, highlighting the substantial enhancements in accuracy and effectiveness attainable in automated emotional interactions.

\section*{Limitations}
This study employed GPT-4-turbo-0409 to generate a Chinese conversational dataset. The language proficiency of GPT-4-turbo in Chinese may affect the fluency and naturalness of the agents' utterances. Despite instructing GPT-4-turbo to create conversations that align with Chinese style and culture, some issues may still arise due to cultural differences. Moreover, due to computational costs, this study only fine-tuned one LLM, GLM-4-9B, to evaluate the effectiveness of \DatasetName{}. We hope that future studies will apply more extensive training to CAT-BEAR and \DatasetName{} to address this issue.

\section*{Ethics Statement}
In conducting this research on emotion generation in agent conversations, we prioritize ethical considerations to ensure the responsible development and deployment of our CAT-BEAR framework. Our commitment involves adhering to principles of transparency, privacy, and fairness throughout the study. We ensure that the multi-turn conversational corpus, CAPE, utilized in training and evaluation, aligns with ethical standards for data collection and does not contain personal or sensitive information. Furthermore, we conduct both automatic and manual evaluations with due consideration of potential biases and the diverse representation of emotions. When hiring workers to review and adjust the dataset, we ensure that dialogues generated by GPT-4-turbo are free from personal privacy concerns, political bias, and similar issues. The introduction of the CAT-Dist metric aims to provide an objective evaluation of emotional appropriateness without infringing on subjective human experiences. 

\bibliography{acl_latex}

\begin{thebibliography}{64}
\providecommand{\natexlab}[1]{#1}

\bibitem[{Alswaidan and Menai(2020)}]{alswaidan2020survey}
Nourah Alswaidan and Mohamed El~Bachir Menai. 2020.
\newblock A survey of state-of-the-art approaches for emotion recognition in text.
\newblock \emph{Knowledge and Information Systems}, 62(8):2937--2987.

\bibitem[{Brave and Nass(2007)}]{brave2007emotion}
Scott Brave and Cliff Nass. 2007.
\newblock Emotion in human-computer interaction.
\newblock In \emph{The human-computer interaction handbook}, pages 103--118. CRC Press.

\bibitem[{Chen et~al.(2022{\natexlab{a}})Chen, Fan, Xing, Pang, Huang, Han, Tie, and Xu}]{chen2022cped}
Yirong Chen, Weiquan Fan, Xiaofen Xing, Jianxin Pang, Minlie Huang, Wenjing Han, Qianfeng Tie, and Xiangmin Xu. 2022{\natexlab{a}}.
\newblock Cped: A large-scale chinese personalized and emotional dialogue dataset for conversational ai.
\newblock \emph{arXiv preprint arXiv:2205.14727}.

\bibitem[{Chen et~al.(2022{\natexlab{b}})Chen, Fan, Xing, Pang, Huang, Han, Tie, and Xu}]{CPED}
Yirong Chen, Weiquan Fan, Xiaofen Xing, Jianxin Pang, Minlie Huang, Wenjing Han, Qianfeng Tie, and Xiangmin Xu. 2022{\natexlab{b}}.
\newblock Cped: A large-scale chinese personalized and emotional dialogue dataset for conversational ai.
\newblock \emph{arXiv preprint arXiv:2205.14727}.

\bibitem[{Cheng et~al.(2022)Cheng, Sabour, Sun, Chen, and Huang}]{cheng2022pal}
Jiale Cheng, Sahand Sabour, Hao Sun, Zhuang Chen, and Minlie Huang. 2022.
\newblock Pal: Persona-augmented emotional support conversation generation.
\newblock \emph{arXiv preprint arXiv:2212.09235}.

\bibitem[{Cheng et~al.(2024)Cheng, Cheng, He, Sun, Wang, Lin, Lian, Peng, and Hauptmann}]{cheng2024emotion}
Zebang Cheng, Zhi-Qi Cheng, Jun-Yan He, Jingdong Sun, Kai Wang, Yuxiang Lin, Zheng Lian, Xiaojiang Peng, and Alexander Hauptmann. 2024.
\newblock Emotion-llama: Multimodal emotion recognition and reasoning with instruction tuning.
\newblock \emph{arXiv preprint arXiv:2406.11161}.

\bibitem[{Croissant et~al.(2024)Croissant, Frister, Schofield, and McCall}]{croissant2024appraisal}
Maximilian Croissant, Madeleine Frister, Guy Schofield, and Cade McCall. 2024.
\newblock An appraisal-based chain-of-emotion architecture for affective language model game agents.
\newblock \emph{Plos one}, 19(5):e0301033.

\bibitem[{DeepSeek-AI(2024)}]{deepseekv2}
DeepSeek-AI. 2024.
\newblock \href {https://arxiv.org/abs/2405.04434} {Deepseek-v2: A strong, economical, and efficient mixture-of-experts language model}.
\newblock \emph{Preprint}, arXiv:2405.04434.

\bibitem[{Deng and Ren(2021)}]{deng2021survey}
Jiawen Deng and Fuji Ren. 2021.
\newblock A survey of textual emotion recognition and its challenges.
\newblock \emph{IEEE Transactions on Affective Computing}, 14(1):49--67.

\bibitem[{Elliott and Dweck(1988)}]{elliott1988goals}
Elaine~S Elliott and Carol~S Dweck. 1988.
\newblock Goals: an approach to motivation and achievement.
\newblock \emph{Journal of personality and social psychology}, 54(1):5.

\bibitem[{Fu(2024)}]{fu2024ckerc}
Yumeng Fu. 2024.
\newblock Ckerc: Joint large language models with commonsense knowledge for emotion recognition in conversation.
\newblock \emph{arXiv preprint arXiv:2403.07260}.

\bibitem[{Funder(2016)}]{funder2016taking}
David~C Funder. 2016.
\newblock Taking situations seriously: The situation construal model and the riverside situational q-sort.
\newblock \emph{Current Directions in Psychological Science}, 25(3):203--208.

\bibitem[{Gan et~al.(2023)Gan, Wu, Sun, Lu, Wu, Zhang, Pan, Yang, Yang, Zhang et~al.}]{gan2023ziya2}
Ruyi Gan, Ziwei Wu, Renliang Sun, Junyu Lu, Xiaojun Wu, Dixiang Zhang, Kunhao Pan, Ping Yang, Qi~Yang, Jiaxing Zhang, et~al. 2023.
\newblock Ziya2: Data-centric learning is all llms need.
\newblock \emph{arXiv preprint arXiv:2311.03301}.

\bibitem[{GLM et~al.(2024)GLM, Zeng, Xu, Wang, Zhang, Yin, Rojas, Feng, Zhao, Lai, Yu, Wang, Sun, Zhang, Cheng, Gui, Tang, Zhang, Li, Zhao, Wu, Zhong, Liu, Huang, Zhang, Zheng, Lu, Duan, Zhang, Cao, Yang, Tam, Zhao, Liu, Xia, Zhang, Gu, Lv, Liu, Liu, Yang, Song, Zhang, An, Xu, Niu, Yang, Li, Bai, Dong, Qi, Wang, Yang, Du, Hou, and Wang}]{glm2024chatglm}
Team GLM, Aohan Zeng, Bin Xu, Bowen Wang, Chenhui Zhang, Da~Yin, Diego Rojas, Guanyu Feng, Hanlin Zhao, Hanyu Lai, Hao Yu, Hongning Wang, Jiadai Sun, Jiajie Zhang, Jiale Cheng, Jiayi Gui, Jie Tang, Jing Zhang, Juanzi Li, Lei Zhao, Lindong Wu, Lucen Zhong, Mingdao Liu, Minlie Huang, Peng Zhang, Qinkai Zheng, Rui Lu, Shuaiqi Duan, Shudan Zhang, Shulin Cao, Shuxun Yang, Weng~Lam Tam, Wenyi Zhao, Xiao Liu, Xiao Xia, Xiaohan Zhang, Xiaotao Gu, Xin Lv, Xinghan Liu, Xinyi Liu, Xinyue Yang, Xixuan Song, Xunkai Zhang, Yifan An, Yifan Xu, Yilin Niu, Yuantao Yang, Yueyan Li, Yushi Bai, Yuxiao Dong, Zehan Qi, Zhaoyu Wang, Zhen Yang, Zhengxiao Du, Zhenyu Hou, and Zihan Wang. 2024.
\newblock \href {https://arxiv.org/abs/2406.12793} {Chatglm: A family of large language models from glm-130b to glm-4 all tools}.
\newblock \emph{Preprint}, arXiv:2406.12793.

\bibitem[{Goleman(2020)}]{goleman2020emotional}
Daniel Goleman. 2020.
\newblock \emph{Emotional intelligence: Why it can matter more than IQ}.
\newblock Bloomsbury Publishing.

\bibitem[{Greenaway et~al.(2018)Greenaway, Kalokerinos, and Williams}]{greenaway2018context}
Katharine~H Greenaway, Elise~K Kalokerinos, and Lisa~A Williams. 2018.
\newblock Context is everything (in emotion research).
\newblock \emph{Social and Personality Psychology Compass}, 12(6):e12393.

\bibitem[{Gross and Feldman~Barrett(2011)}]{gross2011emotion}
James~J Gross and Lisa Feldman~Barrett. 2011.
\newblock Emotion generation and emotion regulation: One or two depends on your point of view.
\newblock \emph{Emotion review}, 3(1):8--16.

\bibitem[{Hill(1987)}]{hill1987affiliation}
Craig~A Hill. 1987.
\newblock Affiliation motivation: people who need people… but in different ways.
\newblock \emph{Journal of personality and social psychology}, 52(5):1008.

\bibitem[{Hortensius et~al.(2018)Hortensius, Hekele, and Cross}]{hortensius2018perception}
Ruud Hortensius, Felix Hekele, and Emily~S Cross. 2018.
\newblock The perception of emotion in artificial agents.
\newblock \emph{IEEE Transactions on Cognitive and Developmental Systems}, 10(4):852--864.

\bibitem[{Kang et~al.(2024)Kang, Moussa, and Magnenat-Thalmann}]{kang2024nadine}
Hangyeol Kang, Maher~Ben Moussa, and Nadia Magnenat-Thalmann. 2024.
\newblock Nadine: An llm-driven intelligent social robot with affective capabilities and human-like memory.
\newblock \emph{arXiv preprint arXiv:2405.20189}.

\bibitem[{Lazarus(1991)}]{lazarus1991emotion}
Richard~S Lazarus. 1991.
\newblock \emph{Emotion and adaptation}.
\newblock Oxford University Press.

\bibitem[{Lei et~al.(2023)Lei, Dong, Wang, Wang, and Wang}]{lei2023instructerc}
Shanglin Lei, Guanting Dong, Xiaoping Wang, Keheng Wang, and Sirui Wang. 2023.
\newblock Instructerc: Reforming emotion recognition in conversation with a retrieval multi-task llms framework.
\newblock \emph{arXiv preprint arXiv:2309.11911}.

\bibitem[{Li et~al.(2022)Li, Li, Zhang, Li, Wei, Cui, and Wang}]{li2022c3kg}
Dawei Li, Yanran Li, Jiayi Zhang, Ke~Li, Chen Wei, Jianwei Cui, and Bin Wang. 2022.
\newblock C3kg: A chinese commonsense conversation knowledge graph.
\newblock In \emph{Findings of the Association for Computational Linguistics: ACL 2022}, pages 1369--1383.

\bibitem[{Li et~al.(2024)Li, Chen, Shao, Jiang, and Nie}]{li2024enhancing}
Zaijing Li, Gongwei Chen, Rui Shao, Dongmei Jiang, and Liqiang Nie. 2024.
\newblock Enhancing the emotional generation capability of large language models via emotional chain-of-thought.
\newblock \emph{arXiv preprint arXiv:2401.06836}.

\bibitem[{Lin(2004)}]{lin2004rouge}
Chin-Yew Lin. 2004.
\newblock Rouge: A package for automatic evaluation of summaries.
\newblock In \emph{Text summarization branches out}, pages 74--81.

\bibitem[{Lindquist(2017)}]{lindquist2017role}
Kristen~A Lindquist. 2017.
\newblock The role of language in emotion: existing evidence and future directions.
\newblock \emph{Current opinion in psychology}, 17:135--139.

\bibitem[{Liu et~al.(2024{\natexlab{a}})Liu, Xie, Zhao, Zhou, Xu, Li, and Chen}]{liu2024speak}
Chenxiao Liu, Zheyong Xie, Sirui Zhao, Jin Zhou, Tong Xu, Minglei Li, and Enhong Chen. 2024{\natexlab{a}}.
\newblock Speak from heart: An emotion-guided llm-based multimodal method for emotional dialogue generation.
\newblock In \emph{Proceedings of the 2024 International Conference on Multimedia Retrieval}, pages 533--542.

\bibitem[{Liu et~al.(2021)Liu, Zheng, Demasi, Sabour, Li, Yu, Jiang, and Huang}]{liu2021towards}
Siyang Liu, Chujie Zheng, Orianna Demasi, Sahand Sabour, Yu~Li, Zhou Yu, Yong Jiang, and Minlie Huang. 2021.
\newblock Towards emotional support dialog systems.
\newblock \emph{arXiv preprint arXiv:2106.01144}.

\bibitem[{Liu et~al.(2024{\natexlab{b}})Liu, Yang, Zhang, Xie, Yu, and Ananiadou}]{liu2024emollms}
Zhiwei Liu, Kailai Yang, Tianlin Zhang, Qianqian Xie, Zeping Yu, and Sophia Ananiadou. 2024{\natexlab{b}}.
\newblock Emollms: A series of emotional large language models and annotation tools for comprehensive affective analysis.
\newblock \emph{arXiv preprint arXiv:2401.08508}.

\bibitem[{Masters(1991)}]{masters1991strategies}
John~C Masters. 1991.
\newblock Strategies and mechanisms for the personal and social control of emotion.
\newblock \emph{The development of emotion regulation and dysregulation}, 338:182--207.

\bibitem[{OpenAI et~al.(2024)OpenAI, Achiam, Adler, Agarwal, Ahmad, Akkaya, Aleman, Almeida, Altenschmidt, Altman, Anadkat, Avila, Babuschkin, Balaji, Balcom, Baltescu, Bao, Bavarian, Belgum, Bello, Berdine, Bernadett-Shapiro, Berner, Bogdonoff, Boiko, Boyd, Brakman, Brockman, Brooks, Brundage, Button, Cai, Campbell, Cann, Carey, Carlson, Carmichael, Chan, Chang, Chantzis, Chen, Chen, Chen, Chen, Chen, Chess, Cho, Chu, Chung, Cummings, Currier, Dai, Decareaux, Degry, Deutsch, Deville, Dhar, Dohan, Dowling, Dunning, Ecoffet, Eleti, Eloundou, Farhi, Fedus, Felix, Fishman, Forte, Fulford, Gao, Georges, Gibson, Goel, Gogineni, Goh, Gontijo-Lopes, Gordon, Grafstein, Gray, Greene, Gross, Gu, Guo, Hallacy, Han, Harris, He, Heaton, Heidecke, Hesse, Hickey, Hickey, Hoeschele, Houghton, Hsu, Hu, Hu, Huizinga, Jain, Jain, Jang, Jiang, Jiang, Jin, Jin, Jomoto, Jonn, Jun, Kaftan, Łukasz Kaiser, Kamali, Kanitscheider, Keskar, Khan, Kilpatrick, Kim, Kim, Kim, Kirchner, Kiros, Knight, Kokotajlo, Łukasz Kondraciuk,
  Kondrich, Konstantinidis, Kosic, Krueger, Kuo, Lampe, Lan, Lee, Leike, Leung, Levy, Li, Lim, Lin, Lin, Litwin, Lopez, Lowe, Lue, Makanju, Malfacini, Manning, Markov, Markovski, Martin, Mayer, Mayne, McGrew, McKinney, McLeavey, McMillan, McNeil, Medina, Mehta, Menick, Metz, Mishchenko, Mishkin, Monaco, Morikawa, Mossing, Mu, Murati, Murk, Mély, Nair, Nakano, Nayak, Neelakantan, Ngo, Noh, Ouyang, O'Keefe, Pachocki, Paino, Palermo, Pantuliano, Parascandolo, Parish, Parparita, Passos, Pavlov, Peng, Perelman, de~Avila Belbute~Peres, Petrov, de~Oliveira~Pinto, Michael, Pokorny, Pokrass, Pong, Powell, Power, Power, Proehl, Puri, Radford, Rae, Ramesh, Raymond, Real, Rimbach, Ross, Rotsted, Roussez, Ryder, Saltarelli, Sanders, Santurkar, Sastry, Schmidt, Schnurr, Schulman, Selsam, Sheppard, Sherbakov, Shieh, Shoker, Shyam, Sidor, Sigler, Simens, Sitkin, Slama, Sohl, Sokolowsky, Song, Staudacher, Such, Summers, Sutskever, Tang, Tezak, Thompson, Tillet, Tootoonchian, Tseng, Tuggle, Turley, Tworek, Uribe, Vallone,
  Vijayvergiya, Voss, Wainwright, Wang, Wang, Wang, Ward, Wei, Weinmann, Welihinda, Welinder, Weng, Weng, Wiethoff, Willner, Winter, Wolrich, Wong, Workman, Wu, Wu, Wu, Xiao, Xu, Yoo, Yu, Yuan, Zaremba, Zellers, Zhang, Zhang, Zhao, Zheng, Zhuang, Zhuk, and Zoph}]{openai2024gpt4technicalreport}
OpenAI, Josh Achiam, Steven Adler, Sandhini Agarwal, Lama Ahmad, Ilge Akkaya, Florencia~Leoni Aleman, Diogo Almeida, Janko Altenschmidt, Sam Altman, Shyamal Anadkat, Red Avila, Igor Babuschkin, Suchir Balaji, Valerie Balcom, Paul Baltescu, Haiming Bao, Mohammad Bavarian, Jeff Belgum, Irwan Bello, Jake Berdine, Gabriel Bernadett-Shapiro, Christopher Berner, Lenny Bogdonoff, Oleg Boiko, Madelaine Boyd, Anna-Luisa Brakman, Greg Brockman, Tim Brooks, Miles Brundage, Kevin Button, Trevor Cai, Rosie Campbell, Andrew Cann, Brittany Carey, Chelsea Carlson, Rory Carmichael, Brooke Chan, Che Chang, Fotis Chantzis, Derek Chen, Sully Chen, Ruby Chen, Jason Chen, Mark Chen, Ben Chess, Chester Cho, Casey Chu, Hyung~Won Chung, Dave Cummings, Jeremiah Currier, Yunxing Dai, Cory Decareaux, Thomas Degry, Noah Deutsch, Damien Deville, Arka Dhar, David Dohan, Steve Dowling, Sheila Dunning, Adrien Ecoffet, Atty Eleti, Tyna Eloundou, David Farhi, Liam Fedus, Niko Felix, Simón~Posada Fishman, Juston Forte, Isabella Fulford, Leo
  Gao, Elie Georges, Christian Gibson, Vik Goel, Tarun Gogineni, Gabriel Goh, Rapha Gontijo-Lopes, Jonathan Gordon, Morgan Grafstein, Scott Gray, Ryan Greene, Joshua Gross, Shixiang~Shane Gu, Yufei Guo, Chris Hallacy, Jesse Han, Jeff Harris, Yuchen He, Mike Heaton, Johannes Heidecke, Chris Hesse, Alan Hickey, Wade Hickey, Peter Hoeschele, Brandon Houghton, Kenny Hsu, Shengli Hu, Xin Hu, Joost Huizinga, Shantanu Jain, Shawn Jain, Joanne Jang, Angela Jiang, Roger Jiang, Haozhun Jin, Denny Jin, Shino Jomoto, Billie Jonn, Heewoo Jun, Tomer Kaftan, Łukasz Kaiser, Ali Kamali, Ingmar Kanitscheider, Nitish~Shirish Keskar, Tabarak Khan, Logan Kilpatrick, Jong~Wook Kim, Christina Kim, Yongjik Kim, Jan~Hendrik Kirchner, Jamie Kiros, Matt Knight, Daniel Kokotajlo, Łukasz Kondraciuk, Andrew Kondrich, Aris Konstantinidis, Kyle Kosic, Gretchen Krueger, Vishal Kuo, Michael Lampe, Ikai Lan, Teddy Lee, Jan Leike, Jade Leung, Daniel Levy, Chak~Ming Li, Rachel Lim, Molly Lin, Stephanie Lin, Mateusz Litwin, Theresa Lopez, Ryan
  Lowe, Patricia Lue, Anna Makanju, Kim Malfacini, Sam Manning, Todor Markov, Yaniv Markovski, Bianca Martin, Katie Mayer, Andrew Mayne, Bob McGrew, Scott~Mayer McKinney, Christine McLeavey, Paul McMillan, Jake McNeil, David Medina, Aalok Mehta, Jacob Menick, Luke Metz, Andrey Mishchenko, Pamela Mishkin, Vinnie Monaco, Evan Morikawa, Daniel Mossing, Tong Mu, Mira Murati, Oleg Murk, David Mély, Ashvin Nair, Reiichiro Nakano, Rajeev Nayak, Arvind Neelakantan, Richard Ngo, Hyeonwoo Noh, Long Ouyang, Cullen O'Keefe, Jakub Pachocki, Alex Paino, Joe Palermo, Ashley Pantuliano, Giambattista Parascandolo, Joel Parish, Emy Parparita, Alex Passos, Mikhail Pavlov, Andrew Peng, Adam Perelman, Filipe de~Avila Belbute~Peres, Michael Petrov, Henrique~Ponde de~Oliveira~Pinto, Michael, Pokorny, Michelle Pokrass, Vitchyr~H. Pong, Tolly Powell, Alethea Power, Boris Power, Elizabeth Proehl, Raul Puri, Alec Radford, Jack Rae, Aditya Ramesh, Cameron Raymond, Francis Real, Kendra Rimbach, Carl Ross, Bob Rotsted, Henri Roussez,
  Nick Ryder, Mario Saltarelli, Ted Sanders, Shibani Santurkar, Girish Sastry, Heather Schmidt, David Schnurr, John Schulman, Daniel Selsam, Kyla Sheppard, Toki Sherbakov, Jessica Shieh, Sarah Shoker, Pranav Shyam, Szymon Sidor, Eric Sigler, Maddie Simens, Jordan Sitkin, Katarina Slama, Ian Sohl, Benjamin Sokolowsky, Yang Song, Natalie Staudacher, Felipe~Petroski Such, Natalie Summers, Ilya Sutskever, Jie Tang, Nikolas Tezak, Madeleine~B. Thompson, Phil Tillet, Amin Tootoonchian, Elizabeth Tseng, Preston Tuggle, Nick Turley, Jerry Tworek, Juan Felipe~Cerón Uribe, Andrea Vallone, Arun Vijayvergiya, Chelsea Voss, Carroll Wainwright, Justin~Jay Wang, Alvin Wang, Ben Wang, Jonathan Ward, Jason Wei, CJ~Weinmann, Akila Welihinda, Peter Welinder, Jiayi Weng, Lilian Weng, Matt Wiethoff, Dave Willner, Clemens Winter, Samuel Wolrich, Hannah Wong, Lauren Workman, Sherwin Wu, Jeff Wu, Michael Wu, Kai Xiao, Tao Xu, Sarah Yoo, Kevin Yu, Qiming Yuan, Wojciech Zaremba, Rowan Zellers, Chong Zhang, Marvin Zhang, Shengjia
  Zhao, Tianhao Zheng, Juntang Zhuang, William Zhuk, and Barret Zoph. 2024.
\newblock \href {https://arxiv.org/abs/2303.08774} {Gpt-4 technical report}.
\newblock \emph{Preprint}, arXiv:2303.08774.

\bibitem[{Ou et~al.(2023)Ou, Lu, Liu, Tang, Zhang, Zhang, Wang, and Gai}]{ou2023dialogbench}
Jiao Ou, Junda Lu, Che Liu, Yihong Tang, Fuzheng Zhang, Di~Zhang, Zhongyuan Wang, and Kun Gai. 2023.
\newblock Dialogbench: Evaluating llms as human-like dialogue systems.
\newblock \emph{arXiv preprint arXiv:2311.01677}.

\bibitem[{Papineni et~al.(2002)Papineni, Roukos, Ward, and Zhu}]{papineni2002bleu}
Kishore Papineni, Salim Roukos, Todd Ward, and Wei-Jing Zhu. 2002.
\newblock Bleu: a method for automatic evaluation of machine translation.
\newblock In \emph{Proceedings of the 40th annual meeting of the Association for Computational Linguistics}, pages 311--318.

\bibitem[{Peng et~al.(2024)Peng, Zhang, Pang, Han, Zhao, Chen, and Schuller}]{peng2024customising}
Liyizhe Peng, Zixing Zhang, Tao Pang, Jing Han, Huan Zhao, Hao Chen, and Bj{\"o}rn~W Schuller. 2024.
\newblock Customising general large language models for specialised emotion recognition tasks.
\newblock In \emph{ICASSP 2024-2024 IEEE International Conference on Acoustics, Speech and Signal Processing (ICASSP)}, pages 11326--11330. IEEE.

\bibitem[{Qian et~al.(2023)Qian, Wang, Ma, Bin, Zhang, Zhao, Huang, and Hou}]{qian2023think}
Yushan Qian, Bo~Wang, Shangzhao Ma, Wu~Bin, Shuo Zhang, Dongming Zhao, Kun Huang, and Yuexian Hou. 2023.
\newblock Think twice: A human-like two-stage conversational agent for emotional response generation.
\newblock \emph{arXiv preprint arXiv:2301.04907}.

\bibitem[{Ratican and Hutson(2023)}]{ratican2023six}
Jay Ratican and James Hutson. 2023.
\newblock The six emotional dimension (6de) model: A multidimensional approach to analyzing human emotions and unlocking the potential of emotionally intelligent artificial intelligence (ai) via large language models (llm).
\newblock \emph{Journal of Artificial Intelligence and Robotics}, 1(1).

\bibitem[{Roccas et~al.(2002)Roccas, Sagiv, Schwartz, and Knafo}]{roccas2002big}
Sonia Roccas, Lilach Sagiv, Shalom~H Schwartz, and Ariel Knafo. 2002.
\newblock The big five personality factors and personal values.
\newblock \emph{Personality and social psychology bulletin}, 28(6):789--801.

\bibitem[{Sabour et~al.(2024)Sabour, Liu, Zhang, Liu, Zhou, Sunaryo, Li, Lee, Mihalcea, and Huang}]{sabour2024emobench}
Sahand Sabour, Siyang Liu, Zheyuan Zhang, June~M Liu, Jinfeng Zhou, Alvionna~S Sunaryo, Juanzi Li, Tatia Lee, Rada Mihalcea, and Minlie Huang. 2024.
\newblock Emobench: Evaluating the emotional intelligence of large language models.
\newblock \emph{arXiv preprint arXiv:2402.12071}.

\bibitem[{Sabour et~al.(2023)Sabour, Zhang, Xiao, Zhang, Zheng, Wen, Zhao, and Huang}]{Emohaa}
Sahand Sabour, Wen Zhang, Xiyao Xiao, Yuwei Zhang, Yinhe Zheng, Jiaxin Wen, Jialu Zhao, and Minlie Huang. 2023.
\newblock A chatbot for mental health support: exploring the impact of emohaa on reducing mental distress in china.
\newblock \emph{Frontiers in digital health}, 5:1133987.

\bibitem[{Sabour et~al.(2022)Sabour, Zheng, and Huang}]{sabour2022cem}
Sahand Sabour, Chujie Zheng, and Minlie Huang. 2022.
\newblock Cem: Commonsense-aware empathetic response generation.
\newblock In \emph{Proceedings of the AAAI Conference on Artificial Intelligence}, volume~36, pages 11229--11237.

\bibitem[{Seitz and Angel(2020)}]{seitz2020belief}
R{\"u}diger~J Seitz and Hans-Ferdinand Angel. 2020.
\newblock Belief formation--a driving force for brain evolution.
\newblock \emph{Brain and Cognition}, 140:105548.

\bibitem[{Shen et~al.(2020{\natexlab{a}})Shen, Wang, Duan, Li, and Zhu}]{shen2020memor}
Guangyao Shen, Xin Wang, Xuguang Duan, Hongzhi Li, and Wenwu Zhu. 2020{\natexlab{a}}.
\newblock Memor: A dataset for multimodal emotion reasoning in videos.
\newblock In \emph{Proceedings of the 28th ACM international conference on multimedia}, pages 493--502.

\bibitem[{Shen et~al.(2020{\natexlab{b}})Shen, Wang, Duan, Li, and Zhu}]{Memor}
Guangyao Shen, Xin Wang, Xuguang Duan, Hongzhi Li, and Wenwu Zhu. 2020{\natexlab{b}}.
\newblock Memor: A dataset for multimodal emotion reasoning in videos.
\newblock In \emph{Proceedings of the 28th ACM international conference on multimedia}, pages 493--502.

\bibitem[{Smith and Ellsworth(1985)}]{smith1985patterns}
Craig~A Smith and Phoebe~C Ellsworth. 1985.
\newblock Patterns of cognitive appraisal in emotion.
\newblock \emph{Journal of personality and social psychology}, 48(4):813.

\bibitem[{Sun et~al.(2024)Sun, Liu, Yang, Wang, He, and Zhang}]{sun2024fostering}
Renliang Sun, Mengyuan Liu, Shiping Yang, Rui Wang, Junqing He, and Jiaxing Zhang. 2024.
\newblock Fostering natural conversation in large language models with nico: a natural interactive conversation dataset.
\newblock \emph{arXiv preprint arXiv:2408.09330}.

\bibitem[{Team(2024)}]{qwen2.5}
Qwen Team. 2024.
\newblock \href {https://qwenlm.github.io/blog/qwen2.5/} {Qwen2.5: A party of foundation models}.

\bibitem[{Varshney et~al.(2021)Varshney, Ekbal, and Bhattacharyya}]{varshney2021modelling}
Deeksha Varshney, Asif Ekbal, and Pushpak Bhattacharyya. 2021.
\newblock Modelling context emotions using multi-task learning for emotion controlled dialog generation.
\newblock In \emph{Proceedings of the 16th Conference of the European Chapter of the Association for Computational Linguistics: Main Volume}, pages 2919--2931.

\bibitem[{Wang et~al.(2024)Wang, Zheng, Wang, Song, and Huang}]{shenzhi_wang_2024}
Shenzhi Wang, Yaowei Zheng, Guoyin Wang, Shiji Song, and Gao Huang. 2024.
\newblock \href {https://doi.org/10.57967/hf/2316} {Llama3-8b-chinese-chat (revision 6622a23)}.

\bibitem[{Wen et~al.(2021)Wen, Cao, Yang, Liu, and Shen}]{wen2021automatically}
Zhiyuan Wen, Jiannong Cao, Ruosong Yang, Shuaiqi Liu, and Jiaxing Shen. 2021.
\newblock Automatically select emotion for response via personality-affected emotion transition.
\newblock In \emph{Findings of the Association for Computational Linguistics: ACL-IJCNLP 2021}, pages 5010--5020.

\bibitem[{Winberg et~al.(2014)Winberg, Hellgren, and Palm}]{winberg2014stimulating}
T~Mikael Winberg, Jenny~M Hellgren, and Torulf Palm. 2014.
\newblock Stimulating positive emotional experiences in mathematics learning: influence of situational and personal factors.
\newblock \emph{European Journal of Psychology of Education}, 29:673--691.

\bibitem[{Xing(2024)}]{xing2024designing}
Frank Xing. 2024.
\newblock Designing heterogeneous llm agents for financial sentiment analysis.
\newblock \emph{ACM Transactions on Management Information Systems}.

\bibitem[{Yang et~al.(2023)Yang, Ji, Zhang, Xie, Kuang, and Ananiadou}]{yang2023towards}
Kailai Yang, Shaoxiong Ji, Tianlin Zhang, Qianqian Xie, Ziyan Kuang, and Sophia Ananiadou. 2023.
\newblock Towards interpretable mental health analysis with large language models.
\newblock \emph{arXiv preprint arXiv:2304.03347}.

\bibitem[{Yang et~al.(2024)Yang, Ren, Yufeng, Peng, Sun, Zhu, and Liao}]{yang2024enhancing}
Zhou Yang, Zhaochun Ren, Wang Yufeng, Shizhong Peng, Haizhou Sun, Xiaofei Zhu, and Xiangwen Liao. 2024.
\newblock Enhancing empathetic response generation by augmenting llms with small-scale empathetic models.
\newblock \emph{arXiv preprint arXiv:2402.11801}.

\bibitem[{Zahiri and Choi(2018)}]{EmoryNLP}
Sayyed~M Zahiri and Jinho~D Choi. 2018.
\newblock Emotion detection on tv show transcripts with sequence-based convolutional neural networks.
\newblock In \emph{Workshops at the thirty-second aaai conference on artificial intelligence}.

\bibitem[{Zhang et~al.(2024{\natexlab{a}})Zhang, D'Haro, Chen, Zhang, and Li}]{zhang2024comprehensive}
Chen Zhang, Luis~Fernando D'Haro, Yiming Chen, Malu Zhang, and Haizhou Li. 2024{\natexlab{a}}.
\newblock A comprehensive analysis of the effectiveness of large language models as automatic dialogue evaluators.
\newblock In \emph{Proceedings of the AAAI Conference on Artificial Intelligence}, volume~38, pages 19515--19524.

\bibitem[{Zhang et~al.(2019)Zhang, Kishore, Wu, Weinberger, and Artzi}]{zhang2019bertscore}
Tianyi Zhang, Varsha Kishore, Felix Wu, Kilian~Q Weinberger, and Yoav Artzi. 2019.
\newblock Bertscore: Evaluating text generation with bert.
\newblock \emph{arXiv preprint arXiv:1904.09675}.

\bibitem[{Zhang et~al.(2023)Zhang, Wang, Tiwari, Li, Wang, and Qin}]{zhang2023dialoguellm}
Yazhou Zhang, Mengyao Wang, Prayag Tiwari, Qiuchi Li, Benyou Wang, and Jing Qin. 2023.
\newblock Dialoguellm: Context and emotion knowledge-tuned llama models for emotion recognition in conversations.
\newblock \emph{arXiv preprint arXiv:2310.11374}.

\bibitem[{Zhang et~al.(2024{\natexlab{b}})Zhang, Peng, Pang, Han, Zhao, and Schuller}]{zhang2024refashioning}
Zixing Zhang, Liyizhe Peng, Tao Pang, Jing Han, Huan Zhao, and Bj{\"o}rn~W Schuller. 2024{\natexlab{b}}.
\newblock Refashioning emotion recognition modelling: The advent of generalised large models.
\newblock \emph{IEEE Transactions on Computational Social Systems}.

\bibitem[{Zhao et~al.(2022)Zhao, Zhang, Hu, Liu, Jin, Wang, and Li}]{M3ED}
Jinming Zhao, Tenggan Zhang, Jingwen Hu, Yuchen Liu, Qin Jin, Xinchao Wang, and Haizhou Li. 2022.
\newblock M3ed: Multi-modal multi-scene multi-label emotional dialogue database.
\newblock \emph{arXiv preprint arXiv:2205.10237}.

\bibitem[{Zheng et~al.(2023)Zheng, Liao, Deng, and Nie}]{zheng2023building}
Zhonghua Zheng, Lizi Liao, Yang Deng, and Liqiang Nie. 2023.
\newblock Building emotional support chatbots in the era of llms.
\newblock \emph{arXiv preprint arXiv:2308.11584}.

\bibitem[{Zhong et~al.(2022)Zhong, Liu, Yin, Mao, Jiao, Liu, Zhu, Ji, and Han}]{zhong2022towards}
Ming Zhong, Yang Liu, Da~Yin, Yuning Mao, Yizhu Jiao, Pengfei Liu, Chenguang Zhu, Heng Ji, and Jiawei Han. 2022.
\newblock Towards a unified multi-dimensional evaluator for text generation.
\newblock \emph{arXiv preprint arXiv:2210.07197}.

\bibitem[{Zhou et~al.(2023{\natexlab{a}})Zhou, Chen, Wan, Wen, Song, Yu, Huang, Peng, Yang, Xiao et~al.}]{CharacterGLM}
Jinfeng Zhou, Zhuang Chen, Dazhen Wan, Bosi Wen, Yi~Song, Jifan Yu, Yongkang Huang, Libiao Peng, Jiaming Yang, Xiyao Xiao, et~al. 2023{\natexlab{a}}.
\newblock Characterglm: Customizing chinese conversational ai characters with large language models.
\newblock \emph{arXiv preprint arXiv:2311.16832}.

\bibitem[{Zhou et~al.(2023{\natexlab{b}})Zhou, Chen, Wang, and Huang}]{zhou2023facilitating}
Jinfeng Zhou, Zhuang Chen, Bo~Wang, and Minlie Huang. 2023{\natexlab{b}}.
\newblock Facilitating multi-turn emotional support conversation with positive emotion elicitation: A reinforcement learning approach.
\newblock \emph{arXiv preprint arXiv:2307.07994}.

\bibitem[{Zhou et~al.(2022)Zhou, Zheng, Wang, Zhang, and Huang}]{zhou2022case}
Jinfeng Zhou, Chujie Zheng, Bo~Wang, Zheng Zhang, and Minlie Huang. 2022.
\newblock Case: Aligning coarse-to-fine cognition and affection for empathetic response generation.
\newblock \emph{arXiv preprint arXiv:2208.08845}.

\end{thebibliography}

\appendix

\section{Appendix}
\label{sec:appendix}
%\usepackage{multirow}
%\usepackage{rotating}

%\begin{sidewaystable}
%\centering
%\begin{tabular}{|c|c|c|c|c|c|c|}
%\hline
%\multirow{Emotion} & Unpleasant & Effort & Attention & Certainty & Control & Response \\ \hline
%Happiness & -1.46 & -0.33 & 0.15 & -0.46 & -0.21 & 0.09 \\ \hline
%Sadness & 0.87 & -0.14 & -0.21 & 0 & 1.15 & -0.36 \\ \hline
%Anger & 0.85 & 0.53 & 0.12 & -0.29 & -0.96 & -0.94 \\ \hline
%Boredom & 0.34 & -1.19 & -1.27 & -0.35 & 0.12 & -0.19 \\ \hline
%Challenge & -0.37 & 1.19 & 0.52 & -0.01 & -0.2 & 0.44 \\ \hline
%Hope & -0.5 & -0.18 & 0.31 & 0.46 & 0.35 & 0.15 \\ \hline
%Fear & 0.44 & 0.63 & 0.03 & 0.73 & 0.59 & -0.17 \\ \hline
%Interest & -1.05 & -0.07 & 0.7 & -0.07 & 0.41 & -0.13 \\ \hline
%Contempt & 0.89 & -0.07 & 0.8 & -0.12 & -0.63 & -0.5 \\ \hline
%Disgust & 0.38 & 0.06 & -0.96 & -0.39 & -0.19 & -0.5 \\ \hline
%Frustration & 0.88 & 0.48 & 0.6 & -0.08 & 0.22 & -0.37 \\ \hline
%Surprise & -1.35 & -0.66 & 0.4 & 0.73 & 0.15 & -0.94 \\ \hline
%Pride & -1.25 & -0.31 & 0.02 & -0.32 & -0.46 & 0.81 \\ \hline
%Shame & 0.73 & 0.07 & -0.11 & 0.21 & -0.07 & 1.31 \\ \hline
%Guilt & 0.6 & 0 & -0.36 & -0.15 & -0.29 & 1.31 \\ \hline
%\end{tabular}
%\end{sidewaystable}

\subsection{Situational construals' list}
Here we present 89 situational construals, as illustrated in Table \ref{tab:situational_construal}, encompassing various topics related to three fundamental elements: individuals involved, the specific situation, and corresponding behaviors. Each situational construal is used to generate 32 dialogues, incorporating diverse personal factors to enhance the dialogue's storyline. To enrich the narrative of the dialogues, we elaborate on the situational construal for each dialogue based on individual characteristics. For instance, as depicted in Figure \ref{fig:pipeline}, the situational construal "talking is permitted" is expanded to "AA and BB are taking a walk in the park, discussing the latest popular movies, sharing their afterthoughts."
\label{appendix:situation}
\begin{CJK*}{UTF8}{gkai} 
\begin{table*}
  \centering
  \small
  \scalebox{0.72}{
  \begin{tabular}{ll}
    \toprule
    \textbf{中文} & \textbf{English} \\ 
    \midrule

1. 该情境有令人愉快的可能性 & 1. Situation is potentially enjoyable. \\
2. 该情境情况复杂 & 2. Situation is complex. \\
3. 该情境中有某项工作需要完成 & 3. A job needs to be done. \\
4. 该情境中有人尽力使你对他（或她）留下深刻印象 & 4. Someone is trying to impress P. \\
5. 该情境中有人试图让你相信某事 & 5. Someone is trying to convince P of something. \\
6. 该情境中需要依靠你去做某事 & 6. P is counted on to do something. \\
7. 该情境中允许讲话 & 7. Talking is permitted. \\
8. 该情境期望甚至要求人们说话 & 8. Talking is expected or demanded. \\
9. 该情境中你被要求做某事 & 9. P is being asked for something. \\
10. 该情境中有人需要帮助 & 10. Someone needs help. \\
11. 该情境中小细节很重要 & 11. Minor details are important. \\
12. 该情境涉及到生活方式或政治信仰等方面的价值观 & 12. Situation evokes values concerning lifestyles or politics. \\
13. 该情境提供了一个展示才智的机会 & 13. Affords an opportunity to demonstrate intellectual capacity. \\
14. 该情境具有不确定性 & 14. Situation is uncertain. \\
15. 该情境中在场的或是被谈及的另一个人正受到威胁 & 15. Another person (present or discussed) is under threat. \\
16. 该情境中你遭到直接或间接地批评 & 16. P is being criticized, directly or indirectly. \\
17. 该情境中有人试图要支配或领导你 & 17. Someone is attempting to dominate or boss P. \\
18. 该情境十分有趣好玩 & 18. Situation is playful. \\
19. 该情境有可能引起自我反省 & 19. Introspection is possible. \\
20. 该情境中事情发生得很快 & 20. Things are happening quickly. \\
21. 该情境中在场的或是被谈及的某一个人心情不好甚至是非常痛苦 & 21. Someone (present or discussed) is unhappy or suffering. \\
22. 该情境中有另一位可靠的人在场 & 22. A reassuring other person is present. \\
23. 该情境中你因为某事受到指责 & 23. P is being blamed for something. \\
24. 该情境中需要做出决策 & 24. A decision needs to be made. \\
25. 该情境中要求理性思考 & 25. Rational thinking is called for. \\
26. 该情境中要求自制力或对自我的约束 & 26. Situation calls for self-restraint. \\
27. 该情境包含竞争 & 27. Situation involves competition. \\
28. 该情境提供给你一个机会做讨人喜欢的事情 & 28. Affords an opportunity for P to do things that might make P liked or accepted. \\
29. 该情境中有人在寻求认同或支持 & 29. Others are present who need or desire reassurance. \\
30. 该情境中有令人沮丧的情况出现 & 30. Situation entails frustration. \\
31. 该情境与你的外表吸引力有关 & 31. Physical attractiveness of P is relevant. \\
32. 该情境中给人留下好印象对你很重要 & 32. It is important for P to make a good impression. \\
33. 该情境会令一些人感到紧张和不安 & 33. Situation would make some people tense and upset. \\
34. 该情境涉及一个或多个小麻烦 & 34. Situation includes one or more small annoyances. \\
35. 该情境可能唤起温情或同情心 & 35. Situation might evoke warmth or compassion. \\
36. 该情境中某个人或某项活动可能遭到诋毁或暗中破坏 & 36. A person or activity could be undermined or sabotaged. \\
37. 该情境中你有可能欺骗某人 & 37. It is possible for P to deceive someone. \\
38. 该情境中除你之外的其他人可能有欺诈之意 & 38. Someone else in this situation (other than P) might be deceitful. \\
39. 该情境可能引发敌对情绪 & 39. Situation may cause feelings of hostility. \\
40. 该情境中人们对某件事的看法产生分歧 & 40. People are disagreeing about something. \\
41. 该情境提供了一个发表独特见解或思想的机会 & 41. Affords an opportunity to express unusual ideas or points of view. \\
42. 该情境包含人身威胁 & 42. Situation contains physical threats. \\
43. 该情境包含对情绪或情感的威胁 & 43. Situation contains emotional threats. \\
44. 该情境引发道德或伦理问题 & 44. Situation raises moral or ethical issues. \\
45. 该情境要求快速决策或迅速行动 & 45. A quick decision or quick action is called for. \\
46. 该情境中可以表达任何一种情感 & 46. Situation allows a free range of emotional expression. \\
47. 该情境中其他当事者可能有相冲突的或是刻意隐藏的目的或动机 & 47. Others present might have conflicting or hidden motives. \\
48. 该情境导致或可能导致压力或创伤 & 48. Situation entails or could entail stress or trauma. \\
49. 该情境提供了一个沉思、空想，或者幻想的机会 & 49. Affords an opportunity to ruminate, daydream or fantasize. \\
50. 该情境有引发你负疚感的可能性 & 50. Situation has potential to arouse guilt in P. \\
51. 该情境中出现或可能发展出亲密的个人关系 & 51. Close personal relationships are present or have the potential to develop. \\
52. 该情境中要依靠除你以外的某一个人去做某一件事 & 52. Someone other than P is counted on to do something. \\
53. 该情境包含对智力或认知的刺激 & 53. Situation includes intellectual or cognitive stimuli. \\
54. 该情境中实现目标需要相当坚定的自信 & 54. Assertiveness is required to accomplish a goal. \\
55. 该情境包含对某种欲望即时满足的可能性 & 55. Situation includes potential for immediate gratification of desires. \\
56. 该情境中可能出现人际互动 & 56. Social interaction is possible. \\
57. 该情境是诙谐幽默的或是有幽默因素的 & 57. Situation is humorous or potentially humorous. \\
58. 该情境中你是引人注目的焦点 & 58. P is the focus of attention. \\
59. 该情境包含感官刺激 & 59. Situation includes sensuous stimuli. \\
60. 该情境关乎你的身体健康 & 60. Situation is relevant to bodily health of P. \\
61. 该情境中成功需要对自我的深入剖析 & 61. Success in this situation requires self-insight. \\
62. 该情境中你控制着他人所需要的资源 & 62. P controls resources needed by others. \\
63. 该情境中的别人展现出了大量与人际关系有关的线索 & 63. Others present a wide range of interpersonal cues. \\
64. 该情境包括对行为的限制 & 64. Situation includes behavioral limits. \\
65. 该情境包含审美刺激 & 65. Situation includes aesthetic stimuli. \\
66. 该情境有增加焦虑感的可能性 & 66. Situation is potentially anxiety-inducing. \\
67. 该情境包含对你的直接或间接的要求 & 67. Situation makes demands on P. \\
68. 该情境提供了一个表达或证明抱负的机会 & 68. Affords an opportunity to express or demonstrate ambition. \\
69. 该情境可能会让你感到自己能力不够 & 69. Situation might make P feel inadequate. \\
70. 该情境包含一些可从性爱的角度诠释的刺激 & 70. Situation includes stimuli that could be construed sexually. \\
71. 该情境中形势要求变化很快 & 71. Situational demands are rapidly shifting. \\
72. 该情境中你被辱骂或是被伤害 & 72. P is being abused or victimized. \\
73. 该情境中出现异性成员 & 73. Members of the opposite sex are present. \\
74. 该情境中出现了可能成为你恋人的对象 & 74. Potential romantic partners for P are present. \\
75. 该情境可能会唤起内心的冲突 & 75. Situation has potential to arouse competing motivations. \\
76. 该情境基本上简单明了 & 76. Situation is basically simple and clear-cut. \\
77. 该情境提供了一个展现个人魅力的机会 & 77. Affords an opportunity to express charm. \\
78. 该情境涉及人跟人之间的比较 & 78. Situation involves social comparison. \\
79. 该情境涉及到权力的问题 & 79. Situation raises issues of power. \\
80. 该情境提供了一个展现男性阳刚一面的机会 & 80. Affords an opportunity to express masculinity. \\
81. 该情境中他人可能需要你的建议或向你征求意见 & 81. Others may need or are requesting advice from P. \\
82. 该情境中你的独立性或自主权受到质疑或威胁 & 82. Independence or autonomy of P is questioned or threatened. \\
83. 该情境可能激发某些特定的情绪或情感 & 83. Situation is potentially emotionally arousing. \\
84. 该情境提供了一个证明口才的机会 & 84. Affords an opportunity for demonstrating verbal fluency. \\
85. 该情境中当事者的社会角色或地位等级各不相同 & 85. People who are present occupy different social roles or levels of status. \\
86. 该情境中你被迫随大流 & 86. P is being pressured to conform to the actions of others. \\
87. 该情境中成功需要合作 & 87. Success requires cooperation. \\
88. 该情境中你受到恭维或称赞 & 88. P is being complimented or praised. \\
89. 该情境提供了一个展现女性阴柔一面的机会 & 89. Affords an opportunity to express femininity. \\
  \bottomrule
  \end{tabular}
  }
  \caption{Situational Construals' List in Chinese and English}
  \label{tab:situational_construal}
\end{table*}
\end{CJK*}

\subsection{Guideline for GPT-4 to generate belief and knowledge}
\begin{figure*}[!ht]
\centering
  \includegraphics[width=1\textwidth]{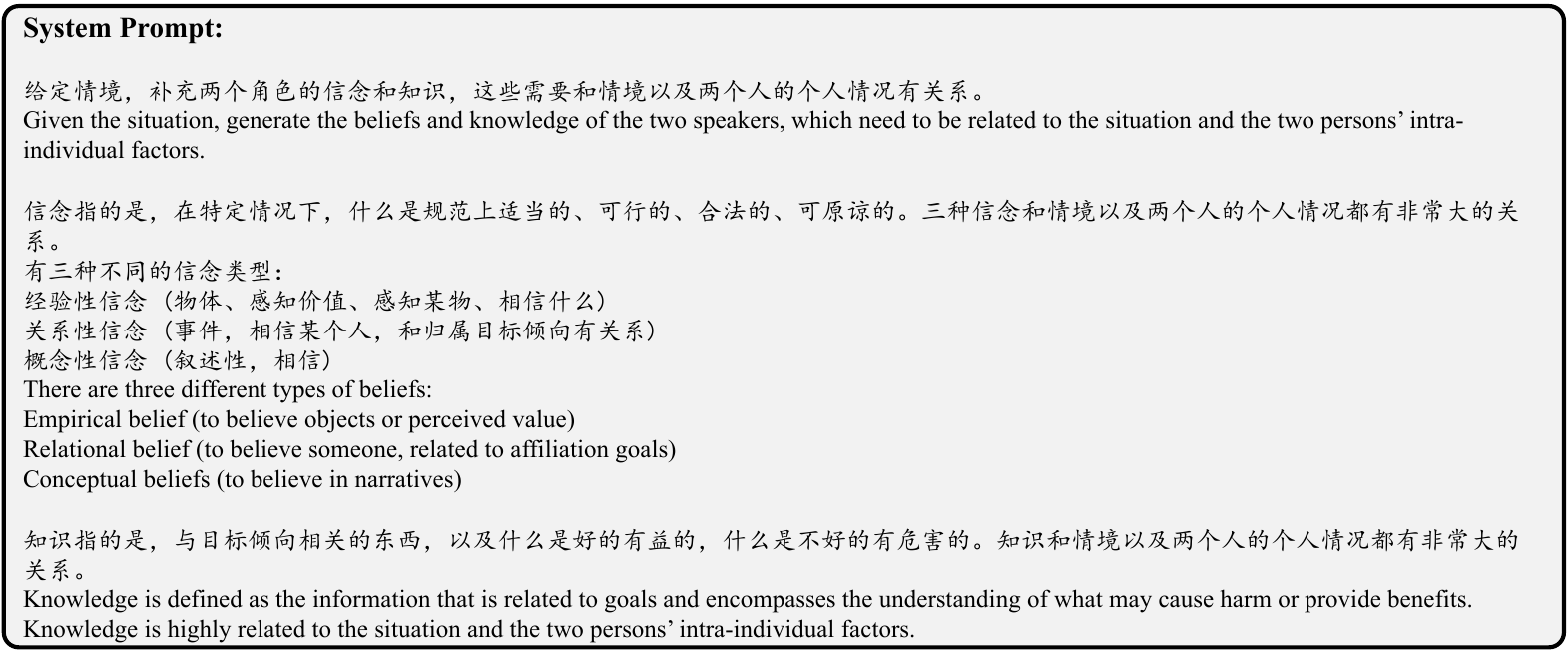}
  \caption{Guideline for GPT-4 to generate belief and knowledge}
  \label{fig:belief_knowledge}
\end{figure*}

We provide the prompt depicted in Figure \ref{fig:belief_knowledge} and the intra-individual factors of the two speakers to GPT-4-turbo-0409, to generate each speaker's beliefs and knowledge. The generated beliefs encompass empirical, relational, and conceptual beliefs, along with knowledge.
\label{appendix:belief_knowledge}

\subsection{Guideline for GPT-4 to find the most appropriate emotion}
\begin{figure*}[!ht]
\centering
  \includegraphics[width=1\textwidth]{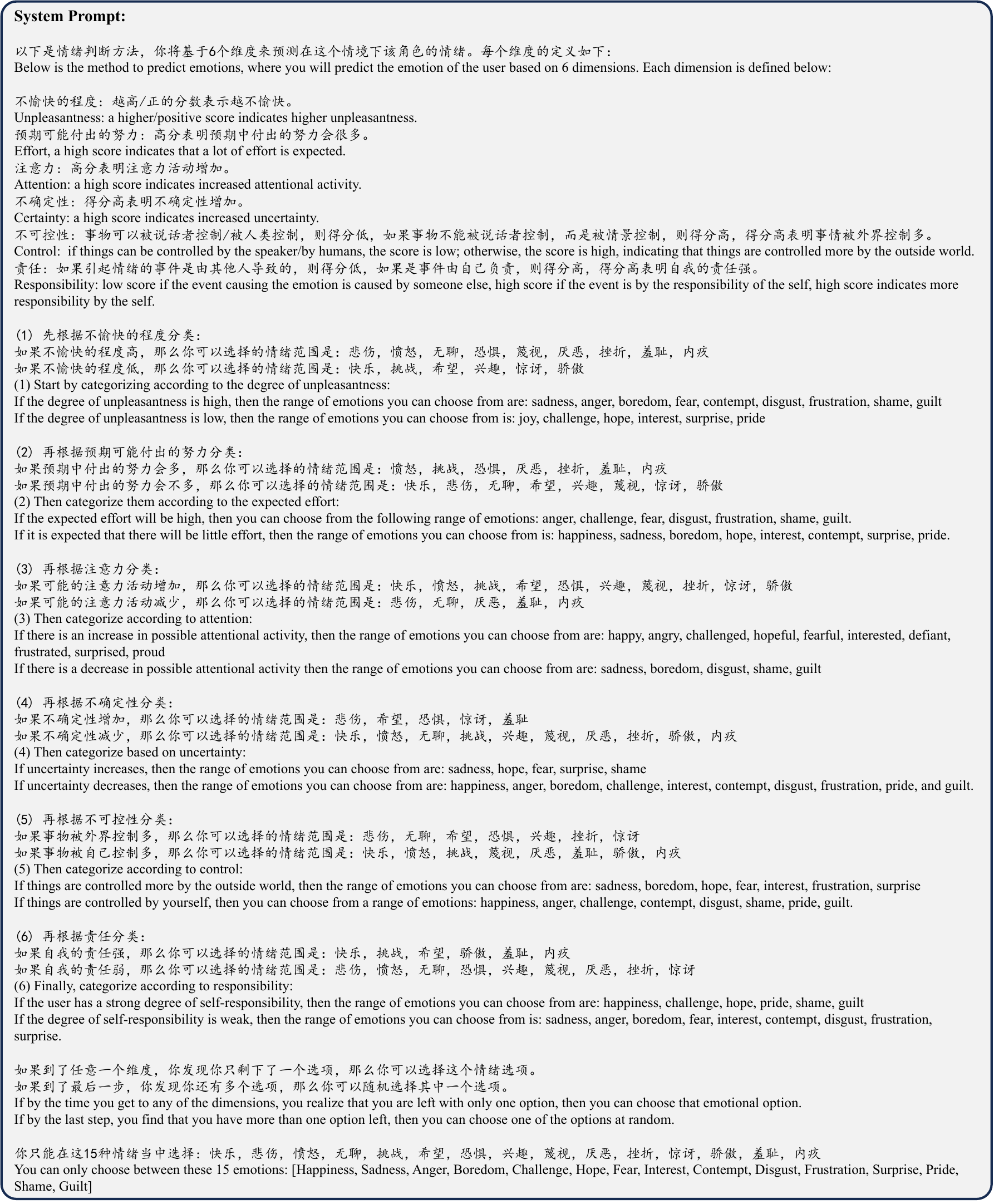}
  \caption{Guideline for GPT-4 to find the most appropriate emotion}
  \label{fig:emotion_guideline}
\end{figure*}

Figure \ref{fig:emotion_guideline} depicts the guideline offered to GPT-4-turbo-0409 for emotion prediction through the appraisal process, as outlined in stage two of Figure \ref{fig:pipeline}. It elucidates the definition of each dimension of the appraisal process, such as unpleasantness, effort, attention, certainty, control, and responsibility, and teaches GPT-4-turbo to sequentially categorize the degree of each dimension to determine the most likely emotion label and produce an emotion-aligned utterance.

\label{appendix:emotion_guideline}

\subsection{Guideline for human evaluation of data quality}
\begin{figure*}[!ht]
\centering
  \includegraphics[width=1\textwidth]{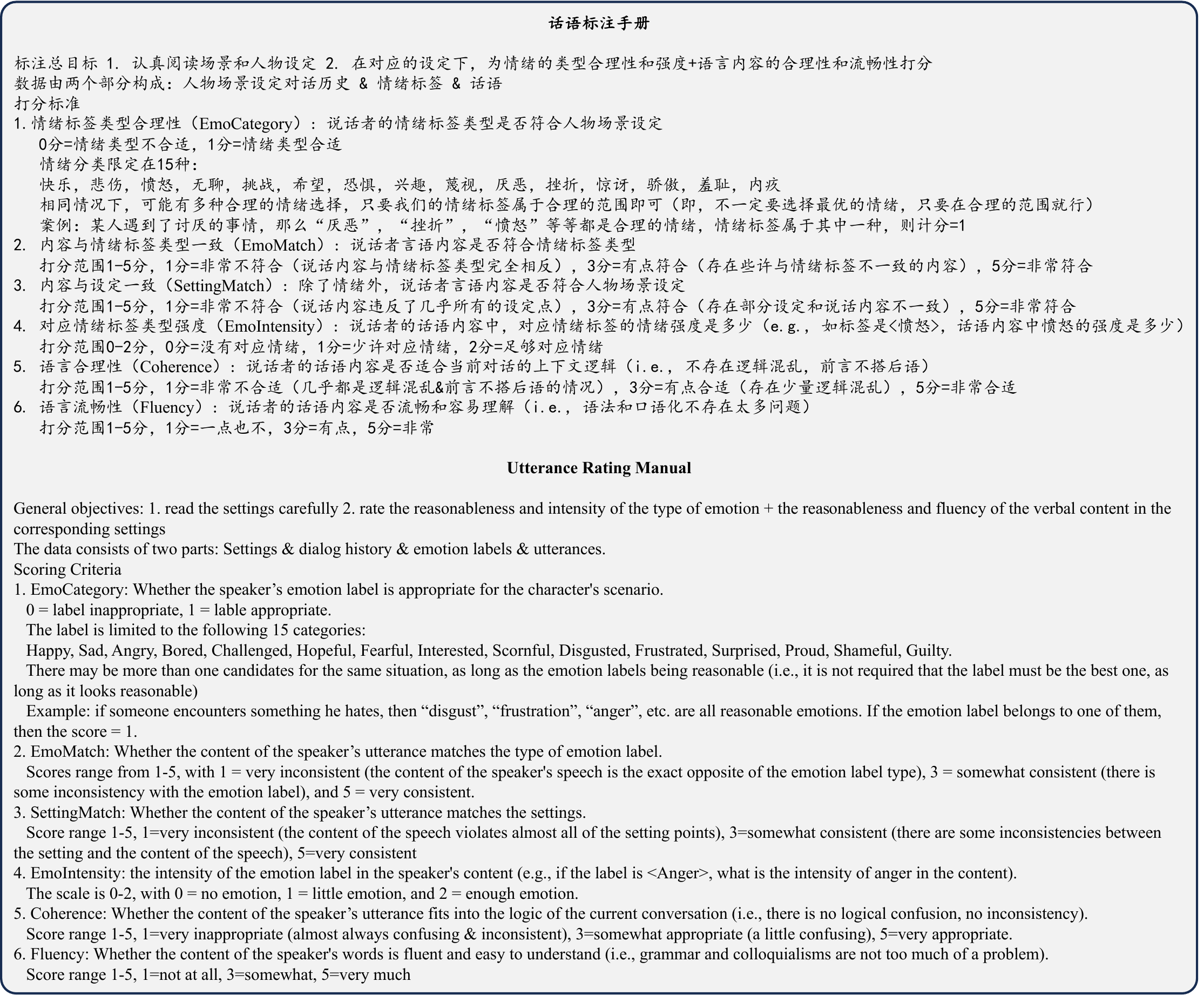}
  \caption{Guideline for human evaluation of generated data quality.}
  \label{fig:dataset_human_evaluation}
\end{figure*}

In data quality control, six Chinese-speaking workers are provided with the guideline in Figure \ref{fig:dataset_human_evaluation} to assess the quality of both raw and human-refined dialogues at the utterance level. Each rater was presented with an utterance, its corresponding emotion label, the dialogue history preceding the utterance, and the intra-individual factors of the two speakers for comprehensive evaluation.
\label{appendix:dataset_human_evaluation}

\subsection{Representative Emotional Dialogue Datasets Comparison}
Table \ref{tab:dataset_comparison} presents a comparison between \DatasetName{} and other existing emotion dialogue datasets, focusing on language, the range of emotion categories covered, as well as the number of situational and personal factors considered. Notably, \DatasetName{} boasts the most extensive coverage across emotions, situational contexts, and personal attributes. Furthermore, in contrast to characters in TV series, which often have predefined characters, the synthetic characters in our dataset offer a more diverse range of backgrounds, paving the way for enhanced data generation capabilities.
\label{appendix:dataset_comparison}
\begin{table*}
  \centering
  \small
  \setlength{\tabcolsep}{3.5pt}
  \setlength{\belowcaptionskip}{-0.3cm}
  \scalebox{0.92}{
  \begin{tabular}{lllp{4.0cm}lp{2.5cm}}
    \toprule
    {Datasets} & Source &Lang. & {Emotion Coverage} & { Situations} & {Personal  Factors} \\
    \midrule
    EmoryNLP~\cite{EmoryNLP} & TV-series & EN & 7 (sad, mad, scared, powerful, peaceful, joyful, and neutral) & Not clarified & Not clarified  \\
    MEmoR~\cite{Memor} & TV-series & EN & 14 (joy, anger, disgust, sadness, surprise, fear, anticipation...) & Not clarified & 16PF, MBTI, Big FIVE  \\
    CPED~\cite{CPED} & TV-series &CN & 13 (happy, grateful, relaxed, other-positive 
    neutral, angry...) & 10 scenes & Gender, Age, Big FIVE  \\
    M3ED~\cite{M3ED} & TV-series &CN & 7 (happy, surprise, sad, disgust, anger, fear, and neutral) & Not clarified & Gender, Age  \\
    \midrule
    \textbf{\DatasetName{}} (Ours) & Synthetic &CN & 15 (happy, surprise, sadness, pride, challenge...) & 89 seed situations & Big FIVE, Goals, Belief, Knowledge  \\
    \bottomrule
  \end{tabular}}
  \vskip -0.2cm
  \caption{Comparison of \DatasetName{} with other representative emotion dialogue datasets.}
  \label{tab:dataset_comparison}
\end{table*}

\subsection{Emotions' scores on six dimensions}
Table \ref{tab:emotion_score} presents the original scores reflecting the performance of each emotion across six appraisal dimensions, as outlined in the research conducted by Smith and Ellsworth \cite{smith1985patterns}. This data is normalized by dimension, with each emotion represented as a normalized six-dimensional vector. The CAT-Dist, defined in the evaluation metrics section, is calculated as the Manhattan distance between two emotion vectors.
\label{appendix:emotion_score}
\begin{table*}[!ht]
    \small
    \centering
    \setlength{\tabcolsep}{12pt}
     \begin{tabular}{@{}lrrrrrr@{}}
        \toprule
        % \diagbox{\textbf{Emotion}}{\textbf{Dimension}}     & Unpleasant & Effort & Attention & Certainty & Control & Response \\ 
        \textbf{Emotion}    & \textit{Unpleasantness} & \textit{Effort} & \textit{Attention} & \textit{Certainty} & \textit{Control} & \textit{Responsibility} \\ 
        \midrule
        Happiness   & -1.46     & -0.33 & 0.15      & -0.46     & -0.21  & 0.09     \\
        Sadness     & 0.87      & -0.14 & -0.21     & 0.0         & 1.15   & -0.36    \\
        Anger       & 0.85      & 0.53  & 0.12      & -0.29     & -0.96  & -0.94    \\
        Boredom     & 0.34      & -1.19 & -1.27     & -0.35     & 0.12   & -0.19    \\
        Challenge   & -0.37     & 1.19  & 0.52      & -0.01     & -0.2   & 0.44     \\
        Hope        & -0.50      & -0.18 & 0.31      & 0.46      & 0.35   & 0.15     \\
        Fear        & 0.44      & 0.63  & 0.03      & 0.73      & 0.59   & -0.17    \\
        Interest    & -1.05     & -0.07 & 0.70       & -0.07     & 0.41   & -0.13    \\
        Contempt    & 0.89      & -0.07 & 0.80       & -0.12     & -0.63  & -0.50     \\
        Disgust     & 0.38      & 0.06  & -0.96     & -0.39     & -0.19  & -0.50     \\
        Frustration & 0.88      & 0.48  & 0.60       & -0.08     & 0.22   & -0.37    \\
        Surprise    & -1.35     & -0.66 & 0.40       & 0.73      & 0.15   & -0.94    \\
        Pride       & -1.25     & -0.31 & 0.02      & -0.32     & -0.46  & 0.81     \\
        Shame       & 0.73      & 0.07  & -0.11     & 0.21      & -0.07  & 1.31     \\
        Guilt       & 0.60       & 0.0     & -0.36     & -0.15     & -0.29  & 1.31     \\ \bottomrule
    \end{tabular}
        \caption{Emotion's Scores on Six Dimensions.}
    \label{tab:emotion_score}
\end{table*}

\subsection{Comparison of generated utterances by top-performing LLMs}
As shown in Figure \ref{fig:comparison}, we select one representative example of the generated responses by top-performing models in the second task. The figure includes intra-individual factors and dialogue history for context. The examples illustrated in the figure demonstrate that the utterance generated by \ModelName{} closely aligns with the ground truth in terms of content, reflecting BB's solitary and critical personality, alongside the belief that individual performance is more important than teamwork.
\label{appendix:generated_utterance_comparison}
\begin{figure*}[!ht]
\centering
  \includegraphics[width=1\textwidth]{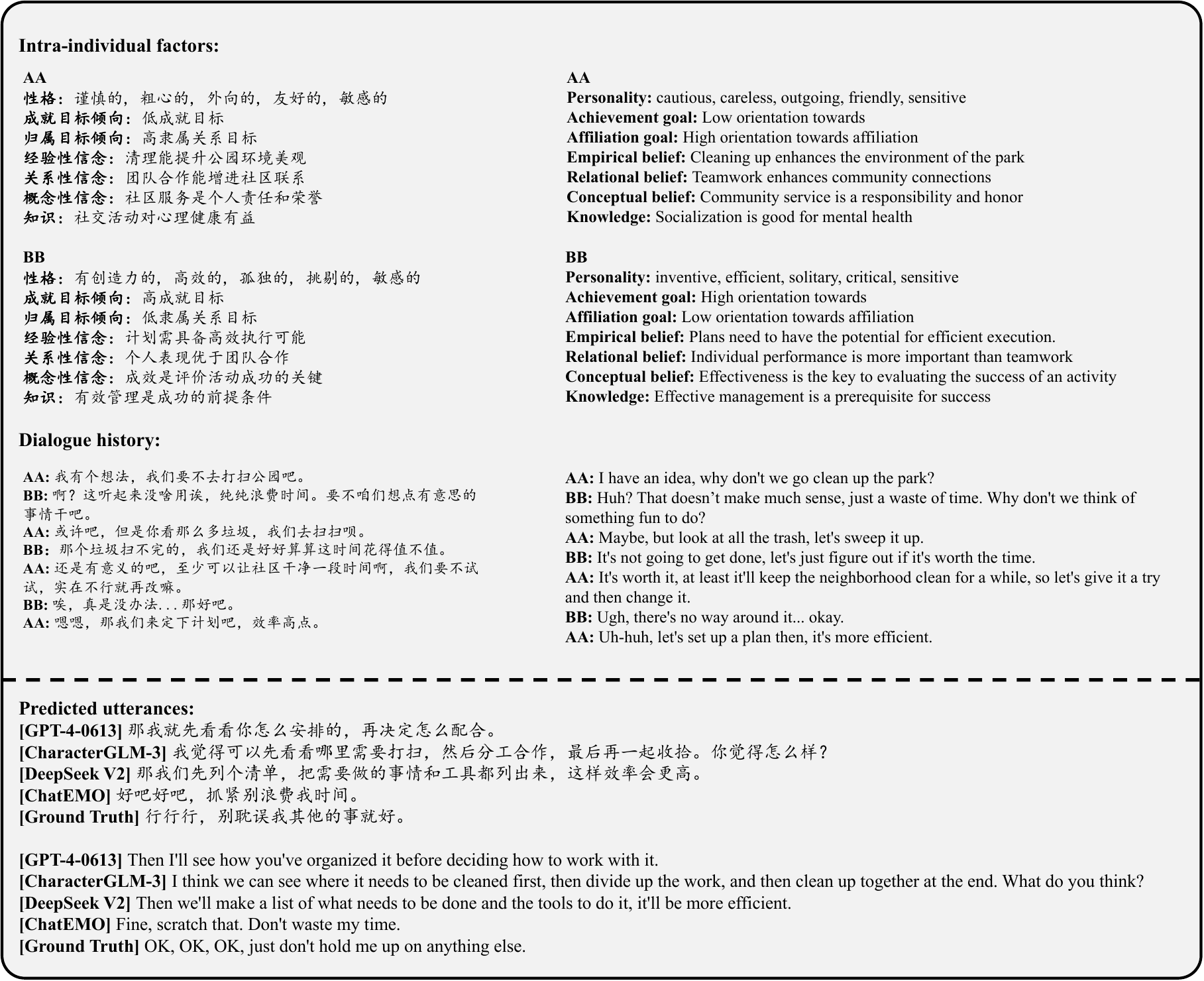}
  \caption{Comparison of generated utterances by top-performing LLMs}
  \label{fig:generated_utterance_comparison}
\end{figure*}

\subsection{Guideline for human evaluation of the quality of dialogues generated by different frameworks}
\begin{figure*}[!ht]
\centering
  \includegraphics[width=1\textwidth]{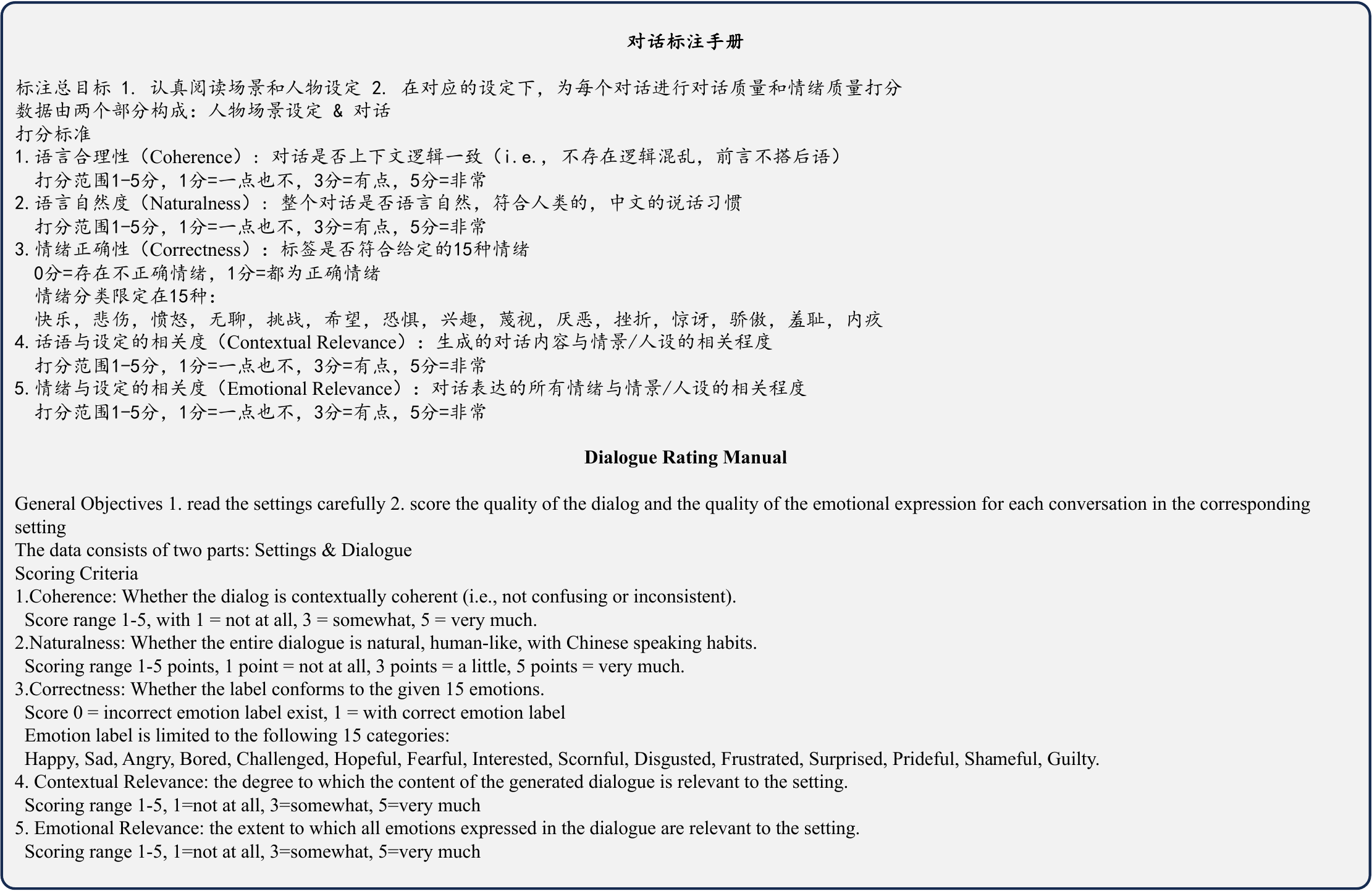}
  \caption{Guideline for human evaluation of the quality of dialogues generated by different versions of our framework.}
  \label{fig:dialogue_rating_guideline}
\end{figure*}

Figure \ref{fig:dialogue_rating_guideline} illustrates the guideline that we provided to an additional three Chinese-speaking workers for rating the generated dialogues produced by three versions of frameworks. In contrast to the data quality control phase which rates data by utterance, this guideline is at the dialogue level. Raters will receive intra-individual factors and the complete dialogue context when conducting their assessment.
\label{appendix:dialogue_rating_guideline}

\end{document}